\documentclass{article}
\usepackage{oat3d-preprint,times}
\usepackage{amsmath,amsfonts,bm}
\usepackage{hyperref}
\usepackage{graphicx,float,booktabs}
\usepackage{colortbl,xcolor}
\usepackage{caption,tabularx}
\usepackage{amssymb}

\title{OAT3D: A Generative Model for 3D Shape- and Resolution-Free Structural Topology Optimization}

\author{\small Hongrui Chen\thanks{These authors contributed equally to this work.
The code is on \href{https://github.com/HongRayChen/OAT3D}{GitHub} and
the dataset is on \href{https://huggingface.co/datasets/ggrayt/oat3d}{Hugging Face}.}\hspace{0.4em},
Kittipong Thiamchaiboonthawee\footnotemark[1]\hspace{0.4em}, Zhen Wei, Amin Heyrani Nobari, Faez Ahmed\\
\fontsize{8.5}{10}\selectfont Department of Mechanical Engineering,
Massachusetts Institute of Technology, Cambridge, MA 02139, USA\\
\small\fontencoding{T1}\ttfamily\{hongrui,kt455,zhen\_wei,ahnobari,faez\}@mit.edu}

\begin{document}

\maketitle

\begin{abstract}
Topology optimization determines how material should be distributed within a design domain to optimize structural performance under prescribed loads, supports, and constraints. The problems are highly nonconvex, and conventional solvers require repeated finite element analyses, making high-resolution 3D topology optimization computationally expensive. Data-driven approaches face an additional bottleneck: generating training datasets requires solving large numbers of these costly problems, so existing work in data-driven topology optimization primarily focuses on 2D problems. Large-scale 3D datasets remain scarce, and existing models remain limited in problem scale and boundary-condition diversity. Here, we introduce Optimize Any Topology-3D (OAT3D), a generative model for high-resolution 3D structural topology optimization. OAT3D introduces a novel sparse problem encoding to represent load and fixity conditions and combines a sparse volumetric variational autoencoder with a conditional diffusion transformer. We construct a dataset of 501,865 optimized topologies, generated using 51,800 GPU-hours and encompassing structures with up to one million elements and substantially more complex loading and support configurations than any existing structural topology optimization dataset. To our knowledge, this is the largest dataset for 3D topology optimization structures. Trained on this dataset, OAT3D generates structures at 512$^3$ resolution and achieves a $44.9\times$ speedup for raw candidate structure generation or $3.65\times$ speedup to reach similar compliance over conventional topology optimization. These contributions expand the scale and diversity of data-driven topology optimization for three-dimensional structural design.
\end{abstract}

\section{Introduction}
\label{sec:oat3d_introduction}
\begin{figure}[H]
    \centering
    \includegraphics[width=\linewidth]{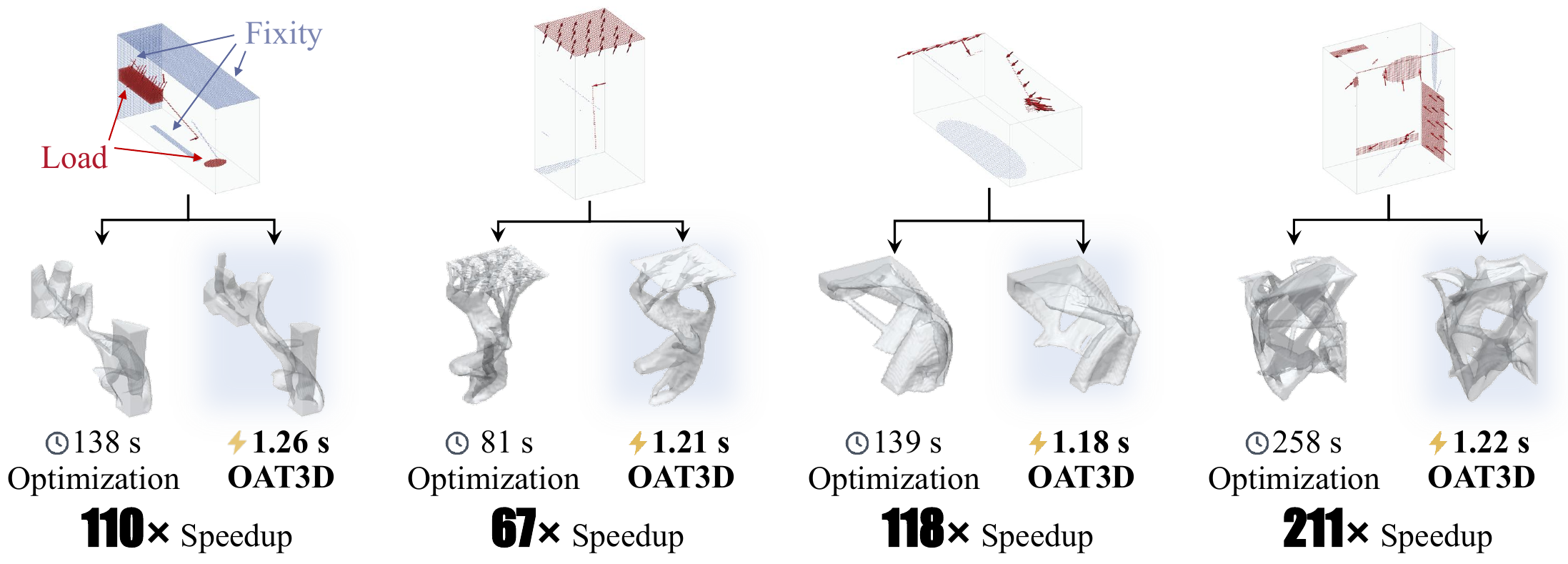}
    \caption{
    Topology optimization finds optimal structures for given loads and fixities, but requires multiple finite element analyses. OAT3D, trained on \textbf{501,865 optimized structures}, can generate a candidate design at $44.9\times$ speedup or reach a comparable compliance with $3.65\times$ speedup.}
    \label{fig:teaserFigure}
\end{figure}
Topology optimization is a physics-based approach to engineering design that determines how material should be distributed within a prescribed domain to optimize for structural performance objectives, with applications in aerospace engineering \citep{zhu2016topology}, civil engineering \citep{baandrup2020closing}, biomedical engineering \citep{sutradhar2010topological}, and beyond. Rather than only adjusting the dimensions or boundaries of a predefined layout, topology optimization allows the arrangement and connectivity of material to change, so that structural members and internal voids emerge through optimization \citep{bendsoe1988generating,bendsoe2004topology}.
A common formulation minimizes compliance, seeking a stiff material layout under prescribed loads and fixities for a given material-volume budget. Density-based methods enable gradient-based optimization \citep{sigmund200199,bendsoe2004topology}.
The design is obtained through an iterative coupling between optimization and physical simulation, which can become computationally demanding as the resolution increases, especially in the three-dimensional setting. Furthermore, full optimization must be rerun for every new loading or fixity configuration.

\begin{figure}[!t]
    \centering
    \includegraphics[width=1\linewidth]{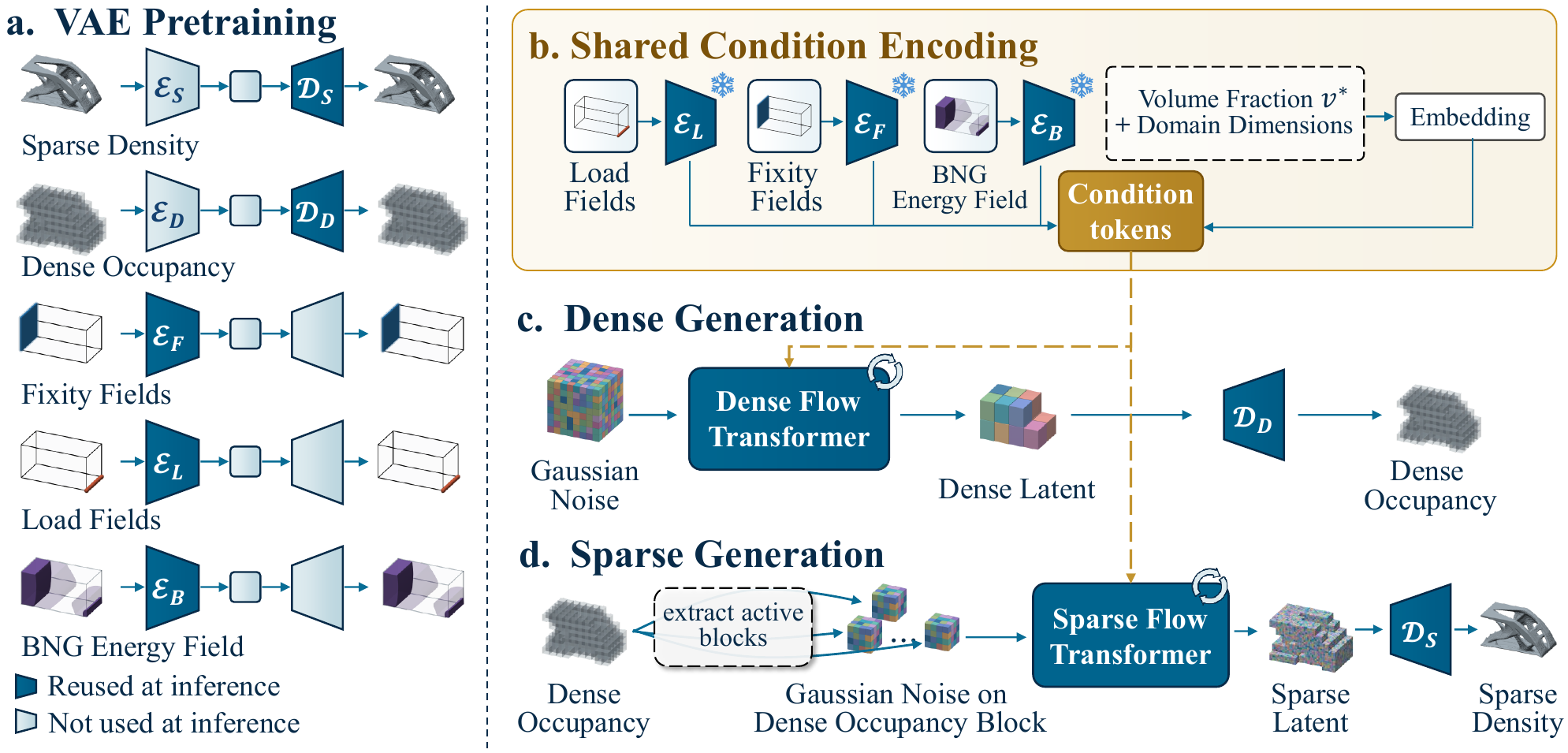}
    \caption{\textbf{OAT3D framework.}
      (a) Variational Autoencoder (VAE)s are pretrained to encode the load and fixity proximity fields and the BNG fields and to represent topology at two resolutions.
      (b) Frozen condition encoders combine the spatial field latents with embeddings of the native domain dimensions and volume fraction to form a shared conditioning sequence.
      (c) Conditioned dense flow matching transforms Gaussian noise into a dense latent, which is decoded into dense occupancy.
      (d) The active dense occupancy cells define the sparse latent generation site, on which conditioned sparse flow matching generates topology latents that are
      decoded into the final density field.}
    \label{fig:framework}

\centering
\scriptsize
\captionof{table}{Comparison of 3D topology optimization datasets. Elements are counted in the finite
element mesh. Aspect ratios are counted as unique combinations of domain dimensions.}
\label{tab:3d-dataset-comparison}
\setlength{\tabcolsep}{3pt}
\renewcommand{\arraystretch}{1.15}
\begin{tabular}{lccccc}
\toprule
Dataset
& \shortstack{Number of\\Topologies $(\uparrow)$}
& \shortstack{Minimum\\Elements $(\uparrow)$}
& \shortstack{Maximum\\Elements $(\uparrow)$}
& \shortstack{Total\\Elements $(\uparrow)$}
& \shortstack{Aspect\\Ratios $(\uparrow)$}\\
\midrule
\cite{keshavarzzadeh2021image} & 2.0K & 0.4K & 3.2K & 3.6M & 1\\
\cite{behzadi2021real} & 2.2K & 8.0K & 128K & 75M & 5\\
SELTO~\citep{dittmer2022selto} & 9.8K & 6.1K & 32K & 315M & 2\\
\cite{herron2024latent} & 66K & \textbf{155K} & 155K & 10.2B & 1\\
NITO-3D~\citep{heyrani2024towards} & 122K & 32K & 48K & 4.3B & 7\\
\textbf{OAT3D dataset (ours)} & \textbf{502K} & 64K & \textbf{1.05M} & \textbf{256B} & \textbf{1.5K}\\
\bottomrule
\end{tabular}\end{figure}

To reduce the cost of solving many distinct design problems, pretrained models learn to predict material layouts from problem conditions rather than running the full iterative topology optimization procedure \citep{yu2019deep,nie2021topologygan,giannone2023aligning,lutheran2025latent}. However, these predictors are largely tied to a fixed discretization. Most 3D models are trained on one or a few grid sizes and aspect ratios (Table~\ref{tab:3d-dataset-comparison}), so a new domain shape or resolution requires retraining a new model with a different size input. This motivates a large-scale generative model that is shape- and resolution-free: a single model, trained on a broad collection of optimized structures, that designs across varied domains, resolutions, and boundary conditions. Since the model must produce topologies that satisfy mechanical constraints, a generative model for optimal structural design requires coupling geometric representation and mechanical conditions. Such a model requires vast training examples that capture material distribution variability across problem specifications, but obtaining a diverse collection of data requires repeatedly solving large linear systems, which limits availability of training data.

We introduce Optimize Any Topology-3D (OAT3D), a generative model that combines spatial mechanical conditioning with two-stage latent flow generation (Figure~\ref{fig:framework}). OAT3D generates designs for any rectangular domain and resolution that fits within the grid, which is what we mean by shape- and resolution-free. The first stage generates dense occupancy, while the second generates continuous material density within the occupied regions. To support learning across varied problems, we construct a collection of 501,865 optimized structures\footnote{We generated approximately 1.2 million topology-optimization solutions before filtering, retaining 501,865 topologies. Successful dataset generation is estimated to have required approximately 51,800 L40S-equivalent GPU-hours, extrapolated from 25,996 logged cases with a mean generation time of 155.43 seconds. } on rectangular domains with randomized loads, fixities, dimensions, and prescribed volume fraction targets. This is the largest, most diverse collection of three-dimensional topology optimized structures. We encode loads and supports as proximity fields, and a Balanced Navier-Green (BNG) field supplements these encodings with an auxiliary description of load-support interaction. The trained model can operate at $512^3$ resolution with a 45$\times$ speedup of raw topology generation or 3.65$\times$ to reach similar compliance compared to direct topology optimization (Figure ~\ref{fig:teaserFigure}). We study the linear elastic minimum-compliance problem and evaluate compliance, material usage, and failure rates for designs from single-generation, best-of-N generation, and warm-started optimization. Lastly, we show that the pretrained model can be reused without retraining for downstream design tasks such as joint-count control and lattice replacement.

Our contributions are as follows:
\begin{itemize}
    \item \textbf{Largest dataset of topology optimized 3D structures.} Our dataset contains 501,865 voxelized structures optimized for minimum compliance (maximum stiffness) with randomized mechanical (load and fixity) conditions, prescribed volume fractions from 0.05 to 0.30, and native meshes containing up to approximately one million elements.
    \item \textbf{Largest generative model for 3D minimum-compliance topology optimization.} We develop a conditional generative framework at $512^3$ resolution with direct density field tailored to structural design problems. As far as we are aware, this is the largest generative model for three-dimensional minimum-compliance structural topology optimization benefiting from an enormous dataset and recent advances in three-dimensional shape generation.
    \item \textbf{Guided generation with spatial encoding of mechanical information.} Combined load, support, and BNG field embeddings supply both generation stages with spatial and mechanical information as guidance.

\end{itemize}

\section{Related work}

\textbf{3D Generative Models.} Recent 3D generation models advance in both geometric representations and generative capabilities.  DreamFusion distills pretrained image diffusion priors into 3D shapes through differentiable rendering \citep{poole2023dreamfusion}. 3DShape2VecSet develops latent-set representations of neural fields \citep{zhang2023shape2vecset}. CLAY and TripoSG develop 3D generation using latent diffusion and rectified flow models \citep{zhang2024clay,li2025triposg}. CAD-oriented shape generation methods such as DeepCAD and BrepGen generate construction sequences or boundary representations that support editability of engineering components \citep{wu2021deepcad,xu2024brepgen}, but do not consider the design's physical performance.

More recently, sparse structured representations have enabled detailed 3D generation without modeling every spatial location at the same granularity. TRELLIS first generates a sparse spatial structure and then local latent features at active locations, which can be decoded into multiple 3D representations \citep{xiang2025structured}. Direct3D-S2 improves sparse volumetric generation through spatial sparse attention, while TRELLIS.2 introduces compact structured latents for both geometry and appearance \citep{wu2025direct3ds2,xiang2026native}. This coarse-to-fine and spatially selective modeling is a good fit for low-volume-fraction topology optimization, where material occupies only part of the design domain and fine-scale modeling can be concentrated within occupied regions. However, the objectives of these visual asset generators focus on geometric fidelity and appearance rather than structural performance under prescribed loads, supports, and material budgets. Although latent diffusion has been explored for 3D structural component generation \citep{herron2024latent}, high-resolution 3D topology generation conditioned on diverse mechanical boundary conditions remains significantly underexplored relative to visual asset generation.

\textbf{Topology Optimization: Dataset and Deep Learning Approaches.} Learning the relationship between mechanical constraints and optimal structure has been explored through deep learning and generative methods~\citep{yu2019deep,nie2021topologygan,maze2023diffusion}, but their generalization to broader domain and resolution sizes are typically limited due to, in part, inherent machine learning architectural limitations, but, mainly, the size, diversity, and availability of topology optimized structure datasets. Table~\ref{tab:3d-dataset-comparison} outlines a survey of the scale and diversity of three-dimensional topology optimized structures in existing literature.

Recent efforts have focused on making these predictors less dependent on fixed discretizations and narrowly prescribed problem settings. NITO-3D \citep{heyrani2024towards} combines a conditional neural implicit density field with an order-invariant point-set representation of loads and supports, enabling prediction at queried coordinates without requiring simulation-derived input fields, and \cite{heyrani2026optimize} took it further with conditional latent diffusion and a neural-field autoencoder trained on completely randomized two-dimensional problems. Its combination of vast training data and conditional generation provides a basis for learning across diverse structural problems. Building on these advances, we aim to develop a unified model for high-resolution three-dimensional minimum-compliance structures across diverse loading and support configurations.

\section{Methodology}
\label{sec:oat3d_method}

\subsection{Three-Dimensional Topology Optimization Dataset Generation}

\label{sec:dataset_generation_body}
We generate three-dimensional topology optimized structures by sampling rectangular domains, loads, supports, and prescribed volume fractions. Each domain contains between $195$k and $3.1$M degrees of freedom. We minimize compliance (maximize stiffness) under linear elastic physics and a volume constraint using Solid Isotropic Material with Penalization (SIMP) method~\citep{bendsoe1989optimal,bendsoe1999material}. Optimization is performed with Projected Gradient Descent (PGD) \citep{heyrani2026pgd}, and finite element analysis is done with GPU-acceleration~\citep{traff2023simple,padhi2023efficient}. A detailed description of the minimum compliance problem formulation and dataset sampling is provided in Appendix~\ref{sec:dataset_generation_appendix} and Appendix~\ref{sec:datasetSampling_appendix}, respectively.

\subsection{Topology and Boundary Condition Representation}
\label{sec:oat3d_representation}

We use a two-stage (dense/sparse) volumetric representation based on TRELLIS and Direct3D-S2~\citep{xiang2025structured,wu2025direct3ds2}. Our generation resolution is limited to $512^3$, while Direct3D-S2 enables up to $1024^3$ resolution.

\textbf{Sparse density VAE} The sparse stage VAE ($\mathcal{E}_S$) encodes the volume from $512^3$ resolution with $8^3$ blocks down to sparse $(512/8)^3 = 64^3$ with $16$ channel latents. Here, sparse means that only $8^3$ blocks with any solid voxels will be encoded. The encoded solid voxels create the dense occupancy that will further be encoded by the dense VAE.

\textbf{Dense occupancy VAE} The dense VAE ($\mathcal{E}_D$) only encodes dense occupancy for the $64^3$ latents provided by the sparse VAE. The occupancy of $64^3$ will be encoded into $16^3$ with 8 channel latents by the dense VAE. This means that the input to the dense VAE is a \textit{single} channel of $64^3$ occupancy instead of 16 from the latent of sparse VAE. The two VAEs can be trained independently: the dense VAE only encodes occupancy, which can be computed directly from the density field without requiring the sparse VAE during training.

Previous work that uses this sparse-dense representation method typically aims to generate surface meshes with no internal structures. In our application, topology optimized structures have complex internals, so we specifically choose to train the sparse VAE to directly represent the continuous densities. Additionally, direct density representation is more efficient in this scenario because it requires fewer occupancy blocks than a signed distance field (SDF), which is the typical method of mesh representation. In fact, we show empirically in Table~\ref{tab:oat3d_support} that using direct density results in over 33.7\% reduction across the mean, median, and 90th percentile quantity in active occupancy blocks.

\paragraph{Boundary condition VAE.}
OAT3D topology generation with diffusion transformer (DiT) requires the information of the boundary condition. Unfortunately, the boundary conditions of topology optimization problems are sparsely populated and spread across multiple channels $\{x,y,z\} \times \{\text{fixity}, +\text{ load direction},-\text{ load direction}\}$. If we were to directly rasterize them into voxels, directly training a VAE solely based on voxel occupancy to capture boundary condition information would be difficult due to the sparsity. Instead, we represent it as a proximity field (outlined in Appendix~\ref{sec:oat3d_conditioning_appendix}). Sparse VAEs $(\mathcal{E}_L,\mathcal{E}_F,\mathcal{E}_B)$ encode these fields into latents that condition the topology optimization generation. Across 2,000 test cases, proximity encoding increases the mean values within retained $8^3$ blocks from $0.0680$ to $0.1561$ for fixity and from $0.00773$ to $0.07761$ for loads, providing denser, stronger reconstruction targets to counteract the background void dominance that can favor near-zero VAE reconstructions, while amplitude normalization alone leaves the spatial sparsity unchanged. We further introduce BNG conditioning to approximate the interaction of loads and supports across the domain (with computation detail in Appendix~\ref{sec:oat3d_conditioning_appendix}). While existing literature uses strain-energy field through a finite element analysis computation \citep{nie2021topologygan, kuszczak2026meta}, our proposed BNG field approximates this without any finite element analysis.

The boundary conditions are then combined with fixed-length embeddings of the domain dimensions and volume fraction. The complete sequence of embeddings conditions the transformers through cross-attention. Additionally, the volume fraction and problem dimension embeddings also condition adaptive layer normalization~\citep{peebles2023scalable}. Appendix~\ref{sec:oat3d_conditioning_appendix} gives the boundary condition conditioning field definitions and encoding procedure. We evaluate reconstruction of the conditioning fields to assess how well their spatial information is preserved in latent space. The mean intersection-over-union (IoU) measures are 0.9926 for fixities, 0.9764 for loads, and 0.9459 for BNG. Appendix~\ref{sec:oat3d_reconstruction_appendix} gives the evaluation thresholds and detailed results.

\subsection{Conditional Generation and Inference}
\label{sec:oat3d_generation}

OAT3D generates structures in two stages (Figure~\ref{fig:framework}). Specifically, the dense DiT starts with a Gaussian noise of $16^3\times 8$ and uses flow matching ~\citep{lipman2023flow} to denoise it into the dense occupancy latent. It then passes into the dense VAE decoder ($\mathcal{D}_D$) to obtain the dense occupancy of $64^3$ (after thresholding the output with 0.3). The dense occupancy resembles a subsampled topology but still lacks the fine detail of the final topology. Next, for each of the occupied dense blocks, we assign a 16 channel Gaussian noise making it sparse $64^3\times16$. This 16 channel latent encodes the fine grained detail of the topology. The sparse DiT uses flow matching to denoise the sparse latent at the same sparse $64^3\times16$ resolution. Finally, the sparse density decoder ($\mathcal{D}_S$) generates the topology within $512^3$ domain.

During both phases of generation, we supply conditioning tokens assembled from the loads, fixities and BNG latents, along with domain dimensions and volume fraction, with cross attention. We train with conditioning dropout enabling classifier-free guidance (CFG)~\citep{ho2022classifier}.  Appendix~\ref{sec:oat3d_generation_appendix} gives the training objective, inference settings, and post-processing procedure.

\section{Experiments}

\subsection{Model training and ablation}
We evaluate reconstruction, BNG conditioning, sampling, and structural performance on 2,000 held out OAT3D problems in the test set. We use 10 flow matching steps with dense CFG 5 and sparse CFG 2.5 with ablation studies in Appendix~\ref{sec:oat3d_sampling_appendix}. Appendix~\ref{sec:oat3d_setup} gives the experimental setup and model configurations. Appendix~\ref{sec:oat3d_reconstruction_appendix} gives the conditioning VAE reconstruction metrics.

\textbf{Effect of density representation.} OAT3D uses density instead of SDF for the sparse VAE representation. The motivation comes from the sparsely populated topology optimized structure. We analyze the number of active blocks between encoding the density field directly versus first setting a threshold of 0.3 then compute a narrow band SDF similar to Direct3D-S2.

\begin{table}[H]
\centering
\captionsetup{font=footnotesize,skip=5pt}

\begin{minipage}[t]{0.40\linewidth}
\vspace{0pt}
\centering
\caption{Active dense occupancy blocks of density vs SDF for test set.}
\label{tab:oat3d_support}
\footnotesize
\setlength{\tabcolsep}{2.5pt}
\renewcommand{\arraystretch}{2.1}
\begin{tabular*}{\linewidth}{@{\extracolsep{\fill}}lrrr@{}}
\toprule
Representation & Mean & Median & \shortstack{90th\\percentile} \\
\midrule
Density & \textbf{516.55} & \textbf{449.5} & \textbf{960} \\
SDF & 806.37 & 734.5 & 1,449 \\
\bottomrule
\end{tabular*}
\end{minipage}\hfill
\begin{minipage}[t]{0.58\linewidth}
\vspace{0pt}
\centering
\caption{Topology VAE reconstruction on 2,000 problems in the test set. Direct3D-S2 models are from their release.}
\label{tab:oat3d_topology_vae}
\footnotesize
\setlength{\tabcolsep}{2.5pt}
\renewcommand{\arraystretch}{1.0}
\begin{tabular*}{\linewidth}{@{\extracolsep{\fill}}lrrr@{}}
\toprule
\shortstack[l]{Representation/\\model} &
\shortstack{Mean\\IoU $\uparrow$} &
\shortstack{Mean\\precision $\uparrow$} &
\shortstack{Mean\\recall $\uparrow$} \\
\midrule
Direct3D-S2, dense & 0.9989 & 0.9993 & 0.9995 \\
OAT3D, dense & \textbf{0.9999} & \textbf{0.9999} & \textbf{1.0000} \\
\midrule
Direct3D-S2, sparse SDF & 0.9531 & \textbf{0.9882} & 0.9634 \\
OAT3D, sparse density & \textbf{0.9762} & 0.9866 & \textbf{0.9892} \\
\bottomrule
\end{tabular*}
\end{minipage}
\end{table}

The result in Table ~\ref{tab:oat3d_support} shows the comparison for the number of active blocks in the dense occupancy. On average direct density representation uses 35.9\% fewer active blocks compared to the SDF representation while retaining the continuous density values without thresholding. We then retrain the dense and sparse VAE based on the density field with results summarized in Table ~\ref{tab:oat3d_topology_vae}. We observe on dense occupancy, OAT3D increases mean IoU from 0.9989 to 0.9999, corresponding to a 90.9\% reduction in residual IoU error. For the sparse representation, OAT3D improves mean IoU from 0.9531 to 0.9762, reducing residual IoU error by 49.3\%.

\textbf{Effect of BNG conditioning.} We formulated a BNG conditioning field that requires no finite element analysis, in place of a strain energy conditioning field that requires an expensive finite element solve. We examine the effect of BNG by comparing models trained with and without BNG conditioning at matched training epochs. Adding BNG increases mean IoU from 0.5958 to 0.6178 for dense occupancy and from 0.7133 to 0.7183 for sparse density generation (Table~\ref{tab:oat3d_bng_comparison}). These results show that the BNG representation of load and support interactions improves generation accuracy, motivating its use in subsequent evaluations.

\textbf{Effect on speedup compared to topology optimization.} The core of this work is to increase the speed of topology optimization with a structural generative model. We compare the computational cost of OAT3D generation with the recorded topology optimization time on the full test set and on the same H200 hardware. Table~\ref{tab:oat3d_cost} shows that mean topology generation time is 1.223s per candidate on an H200 GPU. The dataset was generated on L40S GPUs (86.45s per topology on average). For a same-hardware comparison, we ran topology optimization for the 2,000 test problems in the test set on an H200 (mean optimization time 54.9s). This corresponds to an approximately 44.9$\times$ speedup. To reach a compliance within 5\% (Table \ref{tab:main-overall}), the generated candidate requires 10 steps of post-optimization, which corresponds to a 3.65$\times$ speedup when post-optimization is included.

\begin{table}[H]
\centering
\captionsetup{font=footnotesize,skip=5pt}

\begin{minipage}[t]{0.35\linewidth}
\vspace{0pt}
\centering
\caption{BNG conditioning ablation.}
\label{tab:oat3d_bng_comparison}
\footnotesize
\setlength{\tabcolsep}{2.5pt}
\renewcommand{\arraystretch}{1.0}
\begin{tabular*}{\linewidth}{@{\extracolsep{\fill}}llrr@{}}
\toprule
Stage & BNG  & IoU $\uparrow$ \\
\midrule
Dense & $\times$   & 0.5958 \\
Dense & \checkmark  & \textbf{0.6178} \\
Sparse & $\times$   & 0.7133 \\
Sparse & \checkmark  & \textbf{0.7183} \\
\bottomrule
\end{tabular*}
\end{minipage}\hfill
\begin{minipage}[t]{0.63\linewidth}
\vspace{0pt}
\centering
\caption{Computational cost of OAT3D and dataset.}
\label{tab:oat3d_cost}
\footnotesize
\setlength{\tabcolsep}{2pt}
\renewcommand{\arraystretch}{1.0}
\begin{tabular*}{\linewidth}{@{\extracolsep{\fill}}llrrr@{}}
\toprule
Measurement & Hardware & Cases & Mean & Median \\
\midrule
BNG (ms) & L40S & 2,000 & 36.54 & 31.94 \\
Load voxel filling (ms) & CPU & 2,000 & 19.61 & 4.514 \\
Per candidate (s) & H200 & 2,000 & 1.223 & 1.218 \\
Test generation (s) & H200 & 2,000 & 54.88 & 36.96 \\
Dataset (s) & L40S & 501,865 & 86.45 & 57.94 \\
\bottomrule
\end{tabular*}
\end{minipage}
\end{table}

\begin{table}[H]
\centering
{
\caption{Mean and Median per-problem generation and compliance evaluation time on an H200 GPU for raw generation, best-of-$N$ selection, and post-optimization of the best-of-2 selection, profiled over the 2,000-problem OAT3D test set.}
\label{tab:timing}
\scriptsize
\setlength{\tabcolsep}{3pt}
\begin{tabular}{lrrrrrr}
\toprule
Method & Generation & Compliance & Total & Compliance & \multicolumn{2}{c}{Speedup} \\
\cmidrule(lr){6-7}
 & Time (s) & Evaluation Time (s) & Time (s) & Evaluations & w/ Evaluation & w/o Evaluation \\
\midrule
\multicolumn{7}{l}{\textit{Mean}} \\
Direct Optimization & 54.9 & - & 54.9 & 124 & - & - \\
Raw Generation & 1.22 & 1.67 & 2.90 & 1 & 19.0$\times$ & 44.9$\times$ \\
Best-of-2 & 2.45 & 3.35 & 5.79 & 2 & 9.48$\times$ & 22.4$\times$ \\
Best-of-2 + 5 post-optimization & 2.45 & 8.45 & 10.9 & 7 & 5.04$\times$ & 22.4$\times$ \\
Best-of-2 + 10 post-optimization & 2.45 & 12.6 & 15.0 & 12 & 3.65$\times$ & 22.4$\times$ \\
Best-of-2 + 15 post-optimization & 2.45 & 16.7 & 19.2 & 17 & 2.86$\times$ & 22.4$\times$ \\
\midrule
\multicolumn{7}{l}{\textit{Median}} \\
Direct Optimization & 37.0 & - & 37.0 & 89 & - & - \\
Raw Generation & 1.22 & 1.25 & 2.47 & 1 & 15.0$\times$ & 30.3$\times$ \\
Best-of-2 & 2.44 & 2.50 & 4.94 & 2 & 7.49$\times$ & 15.2$\times$ \\
Best-of-2 + 5 post-optimization & 2.44 & 6.85 & 9.29 & 7 & 3.98$\times$ & 15.2$\times$ \\
Best-of-2 + 10 post-optimization & 2.44 & 10.1 & 12.6 & 12 & 2.94$\times$ & 15.2$\times$ \\
Best-of-2 + 15 post-optimization & 2.44 & 13.5 & 16.0 & 17 & 2.31$\times$ & 15.2$\times$ \\
\bottomrule
\end{tabular}
}
\end{table}

\subsection{Topology Performance Evaluation Against Other Baseline Models}
\label{sec:evaluation}
\label{sec:metrics}
\label{sec:baselines}
\label{sec:selection-refinement}
\textbf{Metrics.} Beyond geometric reconstruction measures, we measure the performance of the generated design: compliance, material usage, and failure for (1) \textbf{raw predictions}, (2) \textbf{selected candidates}, and (3) \textbf{post-optimized designs} under the same problem specifications. These comparisons separate the contributions of generation, candidate selection, and post-optimization.

During evaluation, the input density field is processed as one optimization iteration with the objective evaluated at the end. We do this to ensure consistency between the dataset generation and evaluation of performance and constraint violation.

For test point $i$, the signed compliance error (CE) relative to the reference topology is given by: $\text{CE}_i=\left({C_i-C_i^{\mathrm{ref}}}\right)/{C_i^{\mathrm{ref}}}$. The reference is an optimized design without a guarantee of global optimality, so negative errors remain valid. For element volumes $v_e$, the realized volume fraction and signed relative volume fraction error (VFE) are: $\hat v_i=\left.\left({\sum_e v_e\rho_{i,e}}\right)\middle/{\sum_e v_e}\right.$ and $\text{VFE}_i=\left({\hat v_i-v_i^\star}\right)/{v_i^\star}$,
where $v_i^\star$ is the prescribed volume fraction. Positive VFE denotes excess material. Tables report the mean signed VFE since under-filling naturally leads to higher compliance due to less material presence.

We report mean and median compliance error, mean signed relative volume error, and the fraction of designs that fail at least one numerical, structural, or volume check. A failed numerical solve is possible because a model can output a nonphysical structure that does not satisfy static equilibrium, which is equivalent to an equality constraint violation in the optimization problem. Structural failure denotes a numerically usable evaluation with $C_i>2C_i^{\text{ref}}$. We classify a structure as a volume failure when its volume fraction exceeds the prescribed value by more than $0.10$ (i.e., $\hat{v}_i-v_i^\star>0.10$). This volume criterion is one-sided, so under-filling is permitted. Failed generations are excluded from compliance and volume error summaries.

\textbf{Baselines.} The evaluation set comprises  2,000 problems from the OAT3D test set, spanning varying grid dimensions with both point and distributed loading. Each problem specifies constrained displacement components, nodal force vectors, and a prescribed volume fraction, along with a ground-truth voxelized topology and its corresponding compliance measure as reference targets.

We compare OAT3D with three variants of NITO-3D \citep{heyrani2024towards}: the original model trained on its released dataset, a model retrained on OAT3D dataset, and a parameter-matched model trained on OAT3D dataset. Appendix~\ref{sec:NITO3DBaseline} documents the NITO-3D training procedure.

\textbf{Candidate selection and post-optimization (PO).}
While NITO-3D produces a single deterministic prediction per problem, OAT3D supports inference-time scaling through repeated generative sampling~\citep{heyrani2026optimize}. We evaluate best-of-$N$ sampling for $N\in\{1,2,4,8,16,32\}$ by selecting the raw prediction with the lowest compliance. The selected candidate is used to warm-start topology optimization. We do up to 15 Optimality Criteria (OC) updates~\citep{bendsoe1995optimization,sigmund200199,andreassen2011efficient}.

\subsection{Evaluation Results}

\begin{figure}[H]
    \centering
    \includegraphics[width=\linewidth]{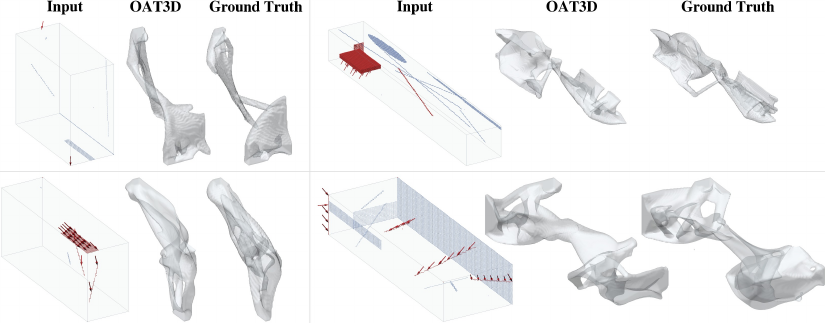}
    \caption{Visualization of (1) The boundary condition input to OAT3D and topology optimization, (2) OAT3D generated topology, (3) Ground truth topology generated from direct optimization. }
    \label{fig:main-input-ground-OAT-post}
\end{figure}

\textbf{Single Generation.}
We select topologies from the test set and visualize them along with the ground truth topologies in Figure ~\ref{fig:main-input-ground-OAT-post}, ~\ref{fig:contact_vlm_3x10_a}, and ~\ref{fig:contact_vlm_3x10_b}. OAT3D has an overall failure rate of 42.40\%, compared with 87.00-99.50\% in the NITO-3D baselines (Table~\ref{tab:main-overall}a). In terms of mechanical performance, it also performs better than all other baselines. Among eligible designs, OAT3D attains lower mean and median CE than the three baselines, at 31.97\% and 23.88\%, respectively. Retraining NITO-3D on OAT3D improves its performance, but increasing its width yields only a small marginal improvement.

The original NITO-3D dataset contains one point load per problem and prescribed volume fractions in the range $[0.15,0.75]$, whereas OAT3D uses a range of $[0.05,0.30]$. We additionally report a comparison in Appendix~\ref{sec:NITO3DBaseline} on the 71 evaluation problems from the OAT3D dataset with exactly one loaded node and a prescribed volume fraction in the shared range. In that setting, OAT3D still significantly outperforms NITO-3D on compliance error and failure measurements across all baselines.

Strikingly, a lower volume fraction range increases the difficulty of the prediction task, demonstrated in a volume-stratification study in Appendix~\ref{sec:volumeStratification}. The failure rate within the volume fraction range of $[0.05,0.10]$ is triple that of $[0.20,0.30]$.

\textbf{Best-of-N.}
\label{sec:best-of-N}
Table~\ref{tab:main-overall}b demonstrates that repeated sampling without retraining improves performance. Increasing the candidate budget from $N=2$ to $N=32$ reduces overall failure from 34.80\% to 19.85\%, while mean and median CE decrease from 30.87\% and 22.80\% to 17.92\% and 11.97\%, respectively. Selection is also accompanied by a shift in mean signed VFE from $-2.18\%$ to $2.17\%$, so the compliance gains should be interpreted alongside the change in material usage. Figure~\ref{fig:bestOf32} in Appendix~\ref{sec:furtherVisualization} illustrates the variation in generated structures for a fixed problem, providing qualitative context for the benefit of candidate selection.

\begin{table}[H]
\centering
\caption{Performance Evaluation Results: (a) Single-generation, (b) Best-of-N selection from OAT3D with and without BNG, (c) Post-optimized from best-of-2 selection.}
\label{tab:main-overall}
\scriptsize
\setlength{\tabcolsep}{3pt}
{
\begin{tabular}{lccrrrr}
\toprule
Method & N & PO Steps & CE mean (\%) $\downarrow$ & CE median (\%) $\downarrow$ & Mean VFE (\%) & Failure (\%) $\downarrow$ \\
\midrule
\multicolumn{7}{l}{\textit{(a) Raw single-generation}}\\
NITO (NITO-trained) & - & - & 43.49 & 41.29 & 27.42 & 99.50 \\
NITO (OAT-trained, 23M) & - & - & 42.59 & 40.99 & 3.75 & 88.65 \\
NITO (OAT-trained, 811M) & - & - & 44.58 & 41.28 & \textbf{0.88} & 87.00 \\
OAT3D & - & - & \textbf{31.97} & \textbf{23.88} & $-$4.37 & \textbf{42.40} \\
OAT3D (no BNG) & - & - & 34.98 & 28.07 & $-$4.43 & 49.85 \\
\midrule
\multicolumn{7}{l}{\textit{(b.1) Best-of-N Selection from OAT3D}}\\
OAT3D & 2 & - & 30.87 & 22.80 & $-$2.18 & 34.80 \\
OAT3D & 4 & - & 26.25 & 18.27 & $-$0.79 & 29.70 \\
OAT3D & 8 & - & 22.74 & 15.99 & \textbf{0.06} & 24.80 \\
OAT3D & 16 & - & 20.23 & 14.30 & 1.06 & 22.15 \\
OAT3D & 32 & - & \textbf{17.92} & \textbf{11.97} & 2.17 & \textbf{19.85} \\
\midrule
\multicolumn{7}{l}{\textit{(b.2) Best-of-$N$ Selection from OAT3D without BNG}} \\
OAT3D (no BNG) & 2 & - & 33.54 & 26.47 & $-$2.83 & 42.10 \\
OAT3D (no BNG) & 4 & - & 28.76 & 21.55 & $-$1.82 & 37.25 \\
OAT3D (no BNG) & 8 & - & 25.27 & 18.92 & $-$0.81 & 33.20 \\
OAT3D (no BNG) & 16 & - & 22.47 & 16.72 & \textbf{$-$0.16} & 29.70 \\
OAT3D (no BNG) & 32 & - & \textbf{20.33} & \textbf{15.04} & 0.84 & \textbf{26.20} \\
\midrule
\multicolumn{7}{l}{\textit{(c) Post-optimized from best-of-2 selection, OC}}\\
OAT3D & 2 & 5 & 17.01 & 12.30 & 0.00 & 18.35 \\
OAT3D & 2 & 10 & 2.71 & 1.42 & 0.00 & 6.45 \\
OAT3D & 2 & 15 & \textbf{1.15} & \textbf{0.50} & 0.00 & \textbf{3.35} \\
\bottomrule
\end{tabular}
}
\end{table}

\textbf{Best-of-N + Post Optimization.}
Post-optimization substantially improves the selected designs (Table ~\ref{tab:main-overall}c). Starting from the best-of-2 prediction, 15 OC updates reduce overall failure from 34.80\% to 3.35\%, with mean and median CE decreasing from 30.87\% and 22.80\% to 1.15\% and 0.50\%, respectively. Much of this improvement is achieved within 10 updates, which yield 6.45\% failure and mean CE of 2.71\%.

\subsection{Application of OAT3D to broader areas of structural optimization}
Structural design often requires modifying an optimized structure to address requirements beyond minimum compliance. Because OAT3D generates by integrating a learned flow, sampling can be steered with external guidance, so the same pretrained generator can be reused for these tasks without retraining or fine-tuning. We demonstrate two engineering applications: joint-count control to adjust structural complexity and lattice replacement based on a selected cell type and member thickness. These applications are conventionally complex direct optimization problems, and their demonstration highlights the value of a large generative model for structural optimization.

\begin{figure}[H]
    \centering
    \includegraphics[width=\linewidth]{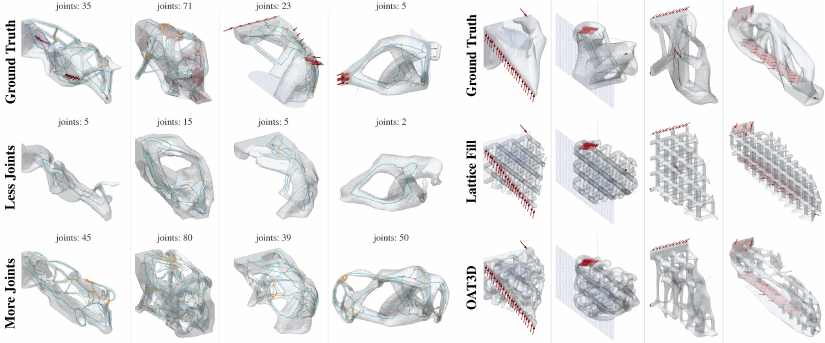}
    \caption{Joint-count control (left) and lattice replacement (right) using OAT3D. Each column shows the reference structure and generated alternatives with fewer and more joints. Orange markers indicate the skeleton junctions used to calculate joint count. Rows show the original structures, direct lattice replacements, and generated results. The first two columns use body-centered cubic lattices; the remaining two use simple cubic lattices.}
    \label{fig:oat3d_joint_control}
\end{figure}

\textbf{Structural characteristics control.}
In density-based topology optimization, structural complexity is typically controlled only implicitly, through the filter radius or length-scale constraints \citep{lazarov2016length}, which alter feature sizes but cannot target a specific characteristic such as the number of joints. Explicit control is difficult because topological quantities such as the number of holes or tunnels are discrete invariants of the design, so existing methods rely on specialized representations, heuristic hole filling, or integer programming \citep{zhang2017explicit, liang2022explicit, he2023hole, zuo2023explicit}. Joint-count control lets a designer request fewer or more branch junctions in the structural skeleton. We adapt TopoCtrl~\citep{chen2026topoctrl} to OAT3D by training separate joint-count regression model on the dense and sparse latents. Gradients of the prediction error guide generation toward the requested count, while a reference-latent penalty encourages similarity to the original structure. Figure~\ref{fig:oat3d_joint_control} shows selected examples with fewer or more joints. Appendix~\ref{sec:oat3d_joint_control_appendix} gives the implementation and more results are in Figure~\ref{fig:joint_appendix14}.

\textbf{Lattice replacement.}
Lattice replacement allows a designer to prescribe cell type and member thickness, but directly substituting a lattice can interrupt the structure because the lattice replacement may miss the load boundary condition completely, resulting in significant increase in compliance. Prior approaches therefore map optimized densities to strut sizes \citep{alzahrani2015design}, optimize porous infill directly at high resolution \citep{wu2017infill}, or use homogenization followed by de-homogenization \citep{groen2018homogenization, wu2021topology}. Maintaining connectivity between adjacent cells and with the surrounding mechanical inputs remains a central challenge. Following TopoEdit~\citep{chen2026topoedit}, we encode the lattice replacement, partially noise its latents, and denoise them under the original loads and supports with guidance toward the edited structure. Figure~\ref{fig:oat3d_joint_control} compares the original structures, direct replacements, and OAT3D outputs. Appendix~\ref{sec:oat3d_lattice_edit_appendix} gives the implementation with more results in Figure ~\ref{fig:lattice_appendix14}.

\section{Discussion and Future Work}
We presented OAT3D, a conditional generative model for three-dimensional minimum-compliance topology optimization. OAT3D combines a dataset of 501,865 optimized structures, a two-stage density representation, and learned boundary-condition encodings augmented by BNG conditioning. We also demonstrate how generation can be guided toward a selected structural characteristic, providing a way to control design features that are difficult to impose directly in conventional density-based optimization.

The present work does have limitations: the dataset does not include prescribed solid or void regions that must remain unchanged during optimization (passive regions). We study linear elastic structures with a single objective. Multi-physics and multi-objective optimization are outside the present scope. Fixity regions act along the global coordinate axes and do not account for rigid kinematic coupling constraints.

Future work will extend the framework to additional design constraints, including requirements for casting, welded assemblies, and sheet-metal fabrication. Developing representations that facilitate conversion of generated density fields into editable CAD models is another direction toward practical structural design. The pretrained model could be extended into other topology optimization problem formulations and physics such as the finite-strain formulation.

\newpage
\section*{AI Use Statement}
We used generative AI tools during the initial exploration of related work and to suggest an initial organization for the plots. AI tools assisted with writing code for training the machine learning models and with deploying training and evaluation code on a compute cluster. We also used them to proofread grammar, suggest table layouts, and generate preliminary drafts of some sections. AI was used to polish and refine writing of this manuscript. The authors take responsibility for the final manuscript, code, and reported results.

\section*{Ethics Statement}
This work involves no human subjects and no personal data. All topology optimization data are generated using computer simulation. The generated topology both from the dataset and the OAT3D model cannot replace real world structural engineering. Even with compliance analysis, the generated design cannot be guaranteed to be manufactured and manufactured design may not satisfy certification without more rigorous validation.

\section*{Reproducibility Statement}

We will release the OAT3D dataset of 501,865 topology-optimized structures and the trained model checkpoint upon paper acceptance. Each example contains the problem specification (domain size, load and fixities, volume fraction), the optimized density field. Every problem is generated with a deterministic 16 bit hexadecimal seed. Together with the sampling procedure in Appendix~\ref{sec:datasetSampling_appendix}, the seed is fully deterministic. We will also release the model evaluation code with all result generated with fixed sampling seeds. The NITO-3D baselines are retrained from the authors' public code and dataset, and Appendix~\ref{sec:NITO3DBaseline} documents the changes for the 23M and 811M variants and compares our reproduction with the published results. Appendix~\ref{sec:oat3d_applications_appendix} gives the regressors, guidance weights, and seeds for the application experiments.

\appendix
\newpage
\section{Topology Optimization and Compliance Minimization Problem Formulation for Dataset Generation}
\label{sec:dataset_generation_appendix}

We use a density-based formulation with Solid Isotropic Material with Penalization (SIMP). The design variables $\boldsymbol{\phi}\in[0,1]^{n_d}$ determine material allocation. They are filtered with a spatial density filter, $\mathbf{H}\left(\boldsymbol{\phi}\right)$, and the resulting densities are assigned to each finite element. We consider compliance minimization subject to mechanical equilibrium and a material-volume constraint:

\begin{equation}
    \label{eq:compliance_problem}
    \begin{aligned}
        \min_{\boldsymbol{\phi}\in[0,1]^{n_d}}
        \quad & C(\boldsymbol{\phi})=\mathbf{f}^{\mathsf{T}}\mathbf{u}\\
        \text{subject to}\quad
        &
        \mathbf{K}\left(\mathbf{H}\left(\boldsymbol{\phi}\right)\right)\mathbf{u}=\mathbf{f}\\
        &
         V\left(\boldsymbol{\phi}\right) \leq v^\star
    \end{aligned}
\end{equation}

where

\begin{equation}
    V\left(\boldsymbol{\phi}\right)=\frac{\sum_{e=1}^{n_d}v_e\phi_e}{\sum_{e=1}^{n_d}v_e}
\end{equation}

Here, $\mathbf{u}$ is the displacement vector for all degrees of freedom in the static system, $\mathbf{f}$ is the assembled force vector, $v_e$ is the volume of element $e$, and $v^\star$ is the prescribed volume fraction. The equilibrium system incorporates zero Dirichlet conditions on the constrained displacement degrees of freedom.

\paragraph{Density filtering and material interpolation.}
The relative densities are computed using a normalized cone filter:
\begin{equation}
    \label{eq:cone_filter}
    \begin{aligned}
        \rho_e=\left[\mathbf{H}\left(\boldsymbol{\phi}\right)\right]_e
        &=\frac{\sum_{j=1}^{n_d} w_{e,j}\phi_j}
                {\sum_{j=1}^{n_d} w_{e,j}} \\
        w_{e,j}
        &=\max\!\left(
        0,1-\frac{\|\mathbf{q}_e-\mathbf{q}_j\|_2}{r_{\min}}
        \right)
    \end{aligned}
\end{equation}
Here, $\mathbf{q}_e$ denotes the center of element $e$, measured in element-width units. We set $r_{\min}=\sqrt{3}$.

We assume isotropic linear elasticity with Young's modulus $E_0=1$ and Poisson's ratio $\nu=1/3$. The filtered densities determine the element stiffness matrices through
\begin{equation}
\label{eq:simp_interpolation}
\mathbf{K}_e(\rho_e)
=\max(\rho_e^p,\epsilon)\,\mathbf{K}_e^{\text{solid}},
\qquad p=3,\qquad \epsilon=10^{-9}
\end{equation}
where $\mathbf{K}_e^{\text{solid}}$ is the stiffness matrix of the solid element. The stiffness floor $\epsilon$ supplies a small residual stiffness to prevent the stiffness matrix from becoming singular. We keep the penalization exponent fixed throughout optimization and apply SIMP directly to the filtered densities.

\paragraph{Optimization.}
We initialize the design uniformly at $\phi_e^{(0)}=v^\star$ and optimize it using pyFANTOM's GPU implementation of projected gradient descent, with adaptive step sizes and conjugate search directions. Each update is projected onto the box and volume constraints. Mechanical equilibrium is solved using multigrid-preconditioned conjugate gradients, with a target relative residual of $10^{-5}$ and at most $400$ solver iterations per solve. The residual is recomputed as

\begin{equation}
\label{eq:equilibrium_residual}
    r_{\mathrm{rel}}
    =\frac{\|\mathbf{f}-\mathbf{K}\mathbf{u}\|_2}{\|\mathbf{f}\|_2}
\end{equation}
Optimization terminates after at most $500$ updates, or when both the following criteria are satisfied:
\begin{equation}
\label{eq:optimization_stopping}
\begin{aligned}
\frac{|C^{(t)}-C^{(t-1)}|}{|C^{(t)}|}&<10^{-5} \\
\frac{\|\boldsymbol{\phi}^{(t)}-\boldsymbol{\phi}^{(t-1)}\|_2}
     {\|\boldsymbol{\phi}^{(t)}\|_2+10^{-12}}&<10^{-2}
\end{aligned}
\end{equation}

\section{Dataset Sampling}
\label{sec:datasetSampling_appendix}
We generate three-dimensional topology optimization problems by randomly sampling the domain dimensions, prescribed volume fraction, loads, and displacement constraints. Each instance has a hexadecimal seed that reproduces its sampled problem configuration. Below, $\mathcal U[a,b]$ denotes a continuous uniform distribution, $\mathcal U\{a,\ldots,b\}$ denotes a uniform distribution over the integers from $a$ to $b$, inclusive, and $\operatorname{Bernoulli}(p)$ returns one with probability $p$.

\subsection{Domain Definition and Discretization}

Each design domain is a cuboid discretized into uniform eight-node hexahedral elements. We determine its dimensions as follows:
\begin{enumerate}
    \item \textbf{Target element count.}
    We sample an element budget
    \begin{equation}
        B\sim\mathcal U\{2^{16},\ldots,2^{20}\}.
    \end{equation}

    \item \textbf{Axis counts.}
    With a minimum of $m=32$ elements along each axis, we sample two
    preliminary dimensions and calculate the third from the remaining
    budget:
    \begin{equation}
        \begin{aligned}
            \widetilde n_x
            &\sim\mathcal U\{m,\ldots,\lfloor B/m^2\rfloor\},\\
            \widetilde n_y\mid\widetilde n_x
            &\sim\mathcal U\{m,\ldots,\lfloor B/(m\widetilde n_x)\rfloor\},\\
            \widetilde n_z
            &=\left\lfloor\frac{B}{\widetilde n_x\widetilde n_y}\right\rfloor.
        \end{aligned}
    \end{equation}
    We round each count upward to a multiple of eight,
    $n_i=8\lceil\widetilde n_i/8\rceil$, for $i\in\{x,y,z\}$.
    If $n_xn_yn_z\notin[2^{16},2^{20}]$, the configuration proposal is
    restarted. Rounding can cause the final element count to differ
    from $B$.

    \item \textbf{Physical dimensions.}
    We normalize the longest dimension to one.
    \begin{align}
        s&=\max(n_x,n_y,n_z)\\
        (L_x,L_y,L_z)&=\frac{(n_x,n_y,n_z)}{s}.
    \end{align}
    Each element is a cube of edge length $1/s$. Geometry is sampled
    in integer node coordinates $\mathbf z=(z_x,z_y,z_z)$, where
    $0\leq z_i\leq n_i$. Physical coordinates can be retrieved with $\mathbf z/s$.

    \item \textbf{Material and volume fraction.}
    We use one isotropic material with solid Young's modulus $E_0=1$
    and Poisson's ratio $\nu=1/3$. The prescribed volume fraction is
    sampled as
    \begin{equation}
        v^\star\sim\mathcal U[0.05,0.30].
    \end{equation}
\end{enumerate}

\subsection{Load and Constraint Specification}

Loads and displacement constraints are assigned to regions of mesh nodes.
A region can consist of a point, line, surface patch, or volume, and may
contain many nodes. We first sample the numbers of regions, then their
locations and geometries, and finally the force vectors or constrained
displacement components.

\begin{enumerate}
    \item \textbf{Numbers of regions.}
    The numbers of load regions and fixity regions are
    \begin{equation}
        N_L\sim\mathcal U\{1,\ldots,10\},
        \qquad N_S\sim\mathcal U\{3,\ldots,10\}.
    \end{equation}
    These count regions, rather than individual loaded nodes or
    constrained degrees of freedom.

    \item \textbf{Interior or exterior placement.}
    An interior region contains only nodes with
    $1\leq z_i\leq n_i-1$ in every coordinate direction.
    An exterior region lies on the domain boundary, where at least one
    coordinate equals zero or the corresponding axis count.
    Each load region is assigned to the interior with probability
    $0.25$. Each fixity region is assigned to the interior with probability $0.10$. Otherwise,
    it is assigned to the exterior.

    \item \textbf{Interior geometry.}
    Conditional on interior placement, the following four types are
    equally likely:
    \begin{itemize}
        \item \textbf{Point:} Sample a node uniformly from the interior,
        with independent coordinates
        $z_i\sim\mathcal U\{1,\ldots,n_i-1\}$.
        \item \textbf{Line:} Sample two distinct interior nodes and
        discretize the segment between them.
        \item \textbf{Surface:} Choose a rectangle or ellipse with equal
        probability. Sample its center coordinates from
        $\mathcal U[1,n_i-1]$ and its normal by normalizing a vector
        with independent components from $\mathcal U[-1,1]$.
        Construct two perpendicular unit vectors in the patch plane
        and independently sample the full lengths along them from
        $\mathcal U[2,\min(n_x,n_y,n_z)/2]$.
        Sample points across the patch and round them to grid nodes and remove duplicates.
        \item \textbf{Volume:} Choose an axis-aligned cuboid or ellipsoid
        with equal probability. Independently sample its center coordinates from $\mathcal U[2,n_i-2]$ and the distance from the center to the edge of the shape along each axis from $\mathcal U[2,0.3n_i]$, then enumerate the grid nodes belonging to the discretized region.
    \end{itemize}
    For interior surfaces and volumes, duplicate nodes are removed and only strictly interior nodes are retained.

    \item \textbf{Exterior geometry.}
    Conditional on exterior placement, points, lines, and surfaces are
    equally likely. Faces are selected uniformly from the six domain
    faces, without weighting by area.
    \begin{itemize}
        \item \textbf{Point:} With probability $0.25$, select one of
        the eight corners uniformly. Otherwise, select a face uniformly
        and then a node uniformly on that face, including its edges
        and corners.
        \item \textbf{Line:} With probability $0.25$, select one of
        the twelve full domain edges uniformly and include all its
        nodes. Otherwise, sample two distinct nodes uniformly on a
        selected face and discretize their segment using the same
        procedure as for interior lines.
        \item \textbf{Surface:} With probability $0.25$, include an
        entire selected face. Otherwise, choose a rectangular or
        elliptical patch with equal probability. Along each in-plane
        axis $i$ of a selected face, sample the center from $\mathcal U[1,n_i-1]$ and
        the half-extent from $\mathcal U[2,n_i/2]$. Construct the patch on the face's integer grid and trim it to the face boundaries.
    \end{itemize}

    \item \textbf{Load directions and magnitudes.}
    For each load region $r$, we sample a vector $\mathbf g_r$
    whose three components are independently drawn from
    $\mathcal U[-0.5,0.5]$. We normalize this vector to obtain
    the unit load direction:
    \begin{equation}
        \mathbf d_r=\frac{\mathbf g_r}{\|\mathbf g_r\|_2}.
    \end{equation}
    We then independently sample a magnitude
    $a_r\sim\mathcal U[0,1]$ for each region and divide it by
    the largest sampled magnitude in the configuration:
    \begin{equation}
        \alpha_r=\frac{a_r}{\max(a_1,\ldots,a_{N_L})}.
    \end{equation}
    Let $m_r=|\mathcal R_r|$ denote the number of nodes in load
    region $\mathcal R_r$. Each node in this region receives
    the force vector $\alpha_r\mathbf d_r/m_r$. Thus, before removing constrained
    components, each region has resultant magnitude $\alpha_r$,
    independent of the node count in a region.

    \item \textbf{Constrained displacement components.}
    Each fixity region receives a mask
    $\mathbf b_r=(b_{r,x},b_{r,y},b_{r,z})$, with independent entries
    \begin{equation}
        b_{r,i}\sim\operatorname{Bernoulli}(0.5).
    \end{equation}
    The mask is shared by all nodes in the region. An entry of one
    imposes a zero Dirichlet condition on that displacement component;
    an entry of zero leaves it unconstrained by this region. A fixity
    node need not have all three degrees of freedom fixed. Overlapping
    fixity regions impose the union of their constraints.
\end{enumerate}
Force components on constrained degrees of freedom are set to zero, and load rows that become entirely zero are removed. At least one nonzero load row must remain. Forces are not renormalized after this step.

\subsection{Dataset Filtering Criteria}
We list the criteria and reason for filtering in Table~\ref{tab:filtering}, along with their effect on the total count of the dataset. We originally performed 1,215,881 optimization runs.
\begin{table}[htbp]
  \centering
  \caption{Filtering criteria applied to the generated dataset.}
  \label{tab:filtering}
  \small
  \renewcommand{\arraystretch}{1.2}
  {
  \begin{tabularx}{\textwidth}{>{\raggedright\arraybackslash}p{3.4cm} >{\raggedright\arraybackslash}X r r}
    \toprule
    \textbf{Criterion for Removal} & \textbf{Reason} & \textbf{Removed} & \textbf{Remaining} \\
    \midrule
    $\geq$ 10 disconnected components after thresholding &
    Because our sampling is random, multiple disconnected islands can form. We decided to filter out cases with 10 or more disconnected components. &
    260,559 & 955,322 \\
    \addlinespace
    No node appears more than once in the load list &
    Early in data generation, loads were aggregated incorrectly. In these cases, the load specification by the sampler did not match the optimization problem that was actually solved. We excluded the affected instances. &
    240,371 & 714,951 \\
    \addlinespace
    Longest domain edge $>$ 512 elements &
    We did not bound the maximum number of elements on any given side in the sampler, but OAT3D training only fits up to 512 elements on any side. &
    134,879 & 580,072 \\
    \addlinespace
    Optimization not converged &
    We removed runs that never exited before running into the 500-iteration cap. &
    56,139 & 523,933 \\
    \addlinespace
    Rigid body motion &
    After thresholding, the topology is checked for rigid body motion. If the topology is not properly physically supported, it is removed. &
    22,068 & 501,865 \\
    \bottomrule
  \end{tabularx}
  }
\end{table}

\section{Topology Representation and Conditional Generation}
\label{sec:oat3d_represetnation_and_conditional_generation}

\subsection{Problem Representation}
\label{sec:oat3d_conditioning_appendix}

\paragraph{Boundary-condition proximity fields.}
We encode boundary conditions (BCs) using three fields for constrained displacement components and six nonnegative fields for signed force directions. For a source at position $\mathbf{x}_i$, the proximity kernel is $k_R(r)=\max(1-r/R,0)$, where $R=5$ native-grid spacings. The fixity proximity field for axis $a$ and load field for signed direction $d$ are
\begin{equation}
b_a(\mathbf{x})=\max_{i\in\mathcal{B}_a} k_R(\|\mathbf{x}-\mathbf{x}_i\|_2),
\qquad
\ell_d(\mathbf{x})=\max_{i\in\mathcal{F}_d} k_R(\|\mathbf{x}-\mathbf{x}_i\|_2)^{p(m_i)}.
\label{eq:oat3d_bc_fields}
\end{equation}
Here, $\mathcal{B}_a$ contains nodes constrained in displacement component $a$, and $\mathcal{F}_d$ contains force components with signed direction $d$. Forces are normalized by the largest absolute component in each problem. The normalized component magnitude $m_i$ is clipped to $[10^{-6},1]$ and sets the exponent $p(m)=3-2.7(m-10^{-6})/(1-10^{-6})$. Larger force values give a slower spatial decay within the same radius. The field value at a source remains one, and an empty source set gives a zero field.

A shared fixity proximity field encoder and a shared load proximity field encoder process active directions separately. We assemble their latents by spatial coordinate and direction to retain the information of each constrained or loaded component. The two boundary condition encoders operate at $513^3$ resolution as the boundary conditions for topology optimization problems are defined at the nodes while the densities are represented at the voxels. The resulting sparse latent resolution is instead $65^3$ with 16 channel latent. The benefit of using sparse VAE to encode the boundary condition is that the number of conditioning tokens required to encode varies with the complexity of the boundary conditions and on average load and fixity require 483.29 tokens to encode.

\paragraph{Balanced Navier-Green conditioning.}
BNG conditioning field serves as the surrogate of strain energy field requiring expensive finite element analysis. A discrete Navier-Green operator, constructed from the zero-mean inverse Neumann Laplacian, maps the balanced load field to an auxiliary displacement field $\widetilde{\mathbf{u}}$. Because the prescribed loads alone are not in equilibrium, we first add the minimum-norm support reactions satisfying the six force and moment balance equations. We calculate the strain-energy-density surrogate
\begin{equation}
W=\mu\,\boldsymbol{\varepsilon}(\widetilde{\mathbf{u}}):
\boldsymbol{\varepsilon}(\widetilde{\mathbf{u}})
+0.5\lambda\bigl[\operatorname{tr}\boldsymbol{\varepsilon}(\widetilde{\mathbf{u}})\bigr]^2,
\label{eq:oat3d_bng_energy}
\end{equation}
where $\lambda$ and $\mu$ are the Lam\'e parameters and $\boldsymbol{\varepsilon}$ is the symmetric discrete displacement gradient. A clipped power transform of the median-normalized energy gives the scalar BNG field. The construction uses the loads and supports without finite element analysis of a candidate structure.

In a prior work, TopoDiff-FF, the authors first used Green's functions~\citep{giannone2023topodiff} as a conditioning field. TopoDiff-FF uses kernel relaxation inspired by Green’s functions to represent loads and boundary conditions. BNG computes approximate support reactions satisfying global force and moment balance, then estimates strain-energy density. We compared raw BNG with a 3D extension of the released TopoDiff-FF kernels against finite element analysis of the fully solid domain. For the kernel comparison, we formed a scalar field using the Euclidean norm of the six channels. Across 200 held-out test set problems, selecting the top 10\% of elements in each field gave a mean IoU with the corresponding FEM energy mask of 0.453 for BNG and 0.172 for TopoDiff-FF.

We encode BNG over the native domain with a sparse VAE, including values below the reconstruction threshold. Each conditioning token packs up to $4^3$ latent sites with 12 channels per site. Table~\ref{tab:oat3d_tokens} gives the token counts. Adding BNG increases sequence length by a mean of 9.81\%, computed from the relative increase for each problem.

\begin{table}[htbp]
\centering
\caption{Conditioning-token counts on 2,000 problems. $L_0$ contains the BC tokens and four scalar tokens; $L_1$ also contains packed BNG tokens.}
\label{tab:oat3d_tokens}
\small
\begin{tabular}{lrrr}
\toprule
Quantity & Mean & Median & 90th percentile \\
\midrule
BC and scalar sequence, $L_0$ & 483.29 & 405 & 913 \\
Added BNG tokens & 37.22 & 32 & 64 \\
Complete sequence, $L_1$ & 520.52 & 442 & 958.3 \\
Added length relative to $L_0$ (\%) & 9.81 & 7.60 & 17.95 \\
\bottomrule
\end{tabular}
\end{table}

\paragraph{Conditioning interface.}
We combine the BC and BNG tokens with four global tokens for the native domain dimensions and volume fraction. The transformers use cross-attention over the complete sequence, and the global tokens also condition adaptive layer normalization~\citep{peebles2023scalable}.

\subsection{Conditional Generation and Inference}
\label{sec:oat3d_generation_appendix}

We train the dense latent generator for dense occupancy and the sparse latent generator for density within active blocks. Sparse training uses occupancy derived from reference densities. At inference, we retain dense cells whose decoded sigmoid probability exceeds 0.3.

Both transformers are trained by flow matching~\citep{lipman2023flow}. For normalized target latent $z_{\mathrm{data}}$, Gaussian noise $\epsilon$, and noise level $\sigma$, the noisy latent and training objective are
\begin{equation}
\begin{aligned}
z_\sigma&=(1-\sigma)z_{\mathrm{data}}+\sigma\epsilon,\\
\mathcal{L}_{\mathrm{flow}}
&=\mathbb{E}\!\left[\left\|u_\theta(z_\sigma,\sigma,c)-(\epsilon-z_{\mathrm{data}})\right\|_2^2/D_z\right],
\end{aligned}
\label{eq:oat3d_flow}
\end{equation}
where $c$ is the encoded problem condition and $D_z$ is the number of latent scalar entries in that problem.

Condition dropout enables classifier-free guidance~\citep{ho2022classifier}. BNG is also dropped whenever fixity-constraint or load information is dropped. The guided velocity is
\begin{equation}
u_{\mathrm{CFG}}=u_\theta(z_\sigma,\sigma,c_{\varnothing})
+w\bigl[u_\theta(z_\sigma,\sigma,c)-u_\theta(z_\sigma,\sigma,c_{\varnothing})\bigr],
\label{eq:oat3d_cfg}
\end{equation}
where $c_{\varnothing}$ is the dropped-condition input. Euler integration runs from noise to data with decreasing $\sigma$.

\paragraph{Load voxel filling.}
After sparse decoding, we apply load voxel filling to the generated density field. For each original load region, a density-based score selects an individual load feature (for example the points, lines, surface patches and volumes), and the corresponding voxels are set to solid.

\newpage
\section{Reconstruction and Conditioning}
\label{sec:oat3d_reconstruction_appendix}

Table~\ref{tab:oat3d_condition_vae} reports reconstruction of the condition fields. Fixity and load proximity field encodings reach mean IoUs of 0.9926 and 0.9764 at a threshold of 0.9. BNG reaches 0.9459 at a threshold of 0.1, with native-domain mean absolute error 0.00550 and root mean squared error 0.00946.

\begin{table}[htbp]
\centering
\caption{Conditioning-field reconstruction on 2,000 problems. BC metrics average active scalar directions within each problem, then average problems. BNG is evaluated over its native domain.}
\label{tab:oat3d_condition_vae}
\small
\begin{tabular}{lrrrr}
\toprule
Field & Threshold & Mean IoU & Mean precision & Mean recall \\
\midrule
Fixity proximity  & $\geq0.9$ & 0.9926 & 0.9947 & 0.9976 \\
Load proximity & $\geq0.9$ & 0.9764 & 0.9864 & 0.9894 \\
BNG & $\geq0.1$ & 0.9459 & 0.9658 & 0.9791 \\
\bottomrule
\end{tabular}
\end{table}

\newpage
\section{Sampling Steps and Classifier-Free Guidance}
\label{sec:oat3d_sampling_appendix}

We vary Euler steps over $\{10,20,30,50\}$ and CFG weights over $\{0.1,1,2.5,5,10,20\}$. Table~\ref{tab:oat3d_dense_grid} shows dense occupancy IoU. At 10 steps, increasing CFG from 5 to 10 changes mean IoU from 0.6178 to 0.6185. Table~\ref{tab:oat3d_sparse_grid} reports sparse material IoU and binary volume-fraction MAE. At 10 steps, sparse CFG 1 gives slightly higher IoU than CFG 2.5 (0.7194 versus 0.7183), while CFG 2.5 gives lower volume-fraction MAE (0.8414 versus 0.8494 percentage points). At sparse CFG 2.5, all larger step counts give lower mean IoU than 10 steps.

\begin{table}[htbp]
\centering
\caption{Dense occupancy IoU with BNG across Euler steps and CFG weights.}
\label{tab:oat3d_dense_grid}
\small
\begin{tabular}{rrrrrrr}
\toprule
Steps / CFG & 0.1 & 1 & 2.5 & 5 & 10 & 20 \\
\midrule
10 & 0.1667 & 0.5815 & 0.6121 & 0.6178 & 0.6185 & 0.6089 \\
20 & 0.1630 & 0.5754 & 0.6088 & 0.6155 & 0.6156 & 0.6082 \\
30 & 0.1632 & 0.5736 & 0.6075 & 0.6149 & 0.6146 & 0.6070 \\
50 & 0.1673 & 0.5717 & 0.6064 & 0.6143 & 0.6141 & 0.6071 \\
\bottomrule
\end{tabular}
\end{table}

\begin{table}[htbp]
\centering
\caption{Sparse generation with BNG across Euler steps and CFG weights. Binary volume-fraction MAE is in percentage points relative to reference volume fraction (VF).}
\label{tab:oat3d_sparse_grid}
\small
\setlength{\tabcolsep}{4pt}
\begin{tabular}{rlrrrrrr}
\toprule
Steps & Metric / CFG & 0.1 & 1 & 2.5 & 5 & 10 & 20 \\
\midrule
10 & IoU $\uparrow$ & 0.6086 & 0.7194 & 0.7183 & 0.6987 & 0.6601 & 0.5372 \\
10 & VF MAE $\downarrow$ & 1.9581 & 0.8494 & 0.8414 & 1.0193 & 1.4274 & 2.2281 \\
\addlinespace
20 & IoU $\uparrow$ & 0.5958 & 0.7086 & 0.7110 & 0.6946 & 0.6658 & 0.6114 \\
20 & VF MAE $\downarrow$ & 2.0368 & 0.8334 & 0.8507 & 1.0170 & 1.4097 & 2.2281 \\
\addlinespace
30 & IoU $\uparrow$ & 0.5913 & 0.7045 & 0.7081 & 0.6925 & 0.6650 & 0.6224 \\
30 & VF MAE $\downarrow$ & 2.0686 & 0.8351 & 0.8557 & 1.0188 & 1.4051 & 2.2368 \\
\addlinespace
50 & IoU $\uparrow$ & 0.5907 & 0.7024 & 0.7059 & 0.6908 & 0.6641 & 0.6253 \\
50 & VF MAE $\downarrow$ & 2.0788 & 0.8378 & 0.8616 & 1.0263 & 1.4243 & 2.2765 \\
\bottomrule
\end{tabular}
\end{table}

\newpage
\section{Experimental Setup}
\label{sec:oat3d_setup}

\paragraph{Data and models.}
Reconstruction and generation comparisons use the first 2,000 reserved problems. We reconstruct with VAE posterior modes and generate with exponential-moving-average (EMA) transformer weights. Dense models are trained for 31 epochs and sparse models for 40 epochs. Unless a parameter is varied, sampling uses 10 Euler steps, scheduler shift 6, dense CFG 5, and sparse CFG 2.5. Generation comparisons use one draw per problem with a fixed seed. Sparse generation comparisons use reference-derived dense occupancy to evaluate.

\paragraph{Reconstruction and generation metrics.}
We report means of IoU, precision, and recall, giving each problem equal weight. Dense occupancy is selected at sigmoid probability $>0.3$ and material occupancy at native density $>0.3$. Fixity and load fields use a localization threshold of $\geq0.9$, and BNG uses $\geq0.1$. The sparse SDF comparison uses its signed zero level. Binary volume-fraction error compares predicted and reference volume fraction:
\begin{equation}
\operatorname{MAE}_{v,\mathrm{bin}}
=\frac{100}{N}\sum_{i=1}^N
\left|\frac{N_i^{\mathrm{pred}}-N_i^{\mathrm{ref}}}{n_{x,i}n_{y,i}n_{z,i}}\right|,
\label{eq:oat3d_binary_vf}
\end{equation}
where $N_i^{\mathrm{pred}}$ and $N_i^{\mathrm{ref}}$ count occupied elements on the native mesh. This error is in percentage points.

\newpage
\section{NITO-3D Baseline Training}
\label{sec:NITO3DBaseline}
We train three NITO-3D baselines following the original training setup of \citet{heyrani2024towards}, except where noted below.

\paragraph{NITO-3D trained on its original dataset.}
Since the original work does not supply model checkpoints and training splits, we retrain NITO-3D using the authors' released dataset and code, and split the data using a random seed. Table~\ref{tab:nito-reproduction} compares our reproduction with the published results.

Exclusively for Table~\ref{tab:nito-reproduction}, we evaluate the performance of the NITO-3D model using NITO-3D's evaluation protocol, which is not identical to the one presented in the main body of this paper. Predictions are binarized at $\rho>0.5$ and evaluated before and after five and ten direct-optimization steps. Problems with $|\mathrm{CE}|\geq1{,}000\%$ are excluded from statistical summaries.

\begin{table}[H]
\centering\small
\caption{NITO-3D reproduction using the authors' public code and an independent test split. Published values are from Table~2 of \citet{heyrani2024towards}. All errors are percentages: CE is signed and VFE is unsigned.}
\label{tab:nito-reproduction}
\begin{tabular}{llrrrr}
\toprule
Stage & Source & CE mean (\%) $\downarrow$ & CE median (\%) $\downarrow$ & $|\mathrm{VFE}|$ mean (\%) $\downarrow$ & $|\mathrm{VFE}|$ median (\%) $\downarrow$ \\
\midrule
Raw & Published & 5.95 & 0.32 & 4.38 & 2.40 \\
Raw & Retrained & 6.552 & 0.414 & 4.393 & 2.755 \\
\arrayrulecolor{black!40}\midrule\arrayrulecolor{black}
5 steps & Published & 0.52 & 0.11 & 0.96 & 0.81 \\
5 steps & Retrained & 0.267 & 0.111 & 0.939 & 0.795 \\
\arrayrulecolor{black!40}\midrule\arrayrulecolor{black}
10 steps & Published & 0.31 & 0.077 & 0.67 & 0.54 \\
10 steps & Retrained & 0.150 & 0.071 & 0.608 & 0.532 \\
\bottomrule
\end{tabular}
\end{table}

\paragraph{NITO-3D trained on OAT3D dataset.}
We train the original architecture, with approximately 23M parameters, on the OAT3D training set with a held out validation set. We select the checkpoint with the lowest validation loss, at epoch 9.

\paragraph{Parameter-matched NITO-3D trained on OAT3D dataset.}
We widen the original architecture without changing its depth, yielding approximately 811M parameters. This model uses the same data and training settings as the 23M baseline, with additional learning-rate warmup and gradient clipping. We select the checkpoint with the lowest validation loss, at epoch 6.

\paragraph{Evaluation on OAT3D dataset.}
Because the original NITO-3D work uses a narrower boundary condition sampling distribution, we additionally evaluate these NITO models on the 71-problem OAT3D subset with exactly one loaded node and a prescribed volume fraction $\in[0.15,0.30]$ using this paper's evaluation protocol.

\begin{table}[H]
\centering\small
\caption{Single-generation performance without candidate selection or refinement on the 71-problem subset with exactly one loaded node and prescribed volume fraction in the nominal shared range of $[0.15,0.30]$. Evaluation process and metric aggregation follow Section~\ref{sec:metrics}.}
\label{tab:sharedloadingNITO}
\setlength{\tabcolsep}{3pt}
\begin{tabular}{lrrrr}
\toprule
Method & CE mean (\%) $\downarrow$ & CE median (\%) $\downarrow$ & Mean VFE (\%) &  Failure (\%) $\downarrow$ \\
\midrule
\multicolumn{5}{l}{\textit{Shared loading and volume conditions: 71 problems}} \\
NITO (NITO-trained) & 25.64 & 24.26 & 11.85 & 94.37 \\
NITO (OAT-trained, 23M) & 26.93 & 18.19 & 7.11 & 39.44 \\
NITO (OAT-trained, 811M) & 25.82 & 19.48 & \textbf{-0.37} & 39.44 \\
OAT3D & \textbf{8.88} & \textbf{3.17} & -4.41 & \textbf{5.63} \\
\bottomrule
\end{tabular}
\end{table}

\paragraph{Findings.}
Our reproduction achieves performance comparable to the published results on the original dataset (Table~\ref{tab:nito-reproduction}). On the full OAT3D evaluation set, retraining reduces overall failure from 99.50\% to 88.65\%, but widening the OAT3D-trained model from approximately 23M to 811M parameters yields only a small marginal reduction to 87.00\%.

The shared-condition comparison in Table~\ref{tab:sharedloadingNITO} reinforces this pattern. On these 71 problems, retraining with the OAT3D dataset reduces overall failure on the OAT3D dataset from 94.37\% to 39.44\%, but widening the model leaves this rate unchanged. OAT3D achieves an overall failure rate of 5.63\%, with mean and median CE of 8.88\% and 3.17\%, respectively. Within the overlapping load-count and volume-fraction conditions, OAT still retains a performance advantage.

Together, these results suggest that width scaling alone, despite a more than 35-fold increase in parameters, does not close the performance gap.

\newpage
\section{Further Visualization of Results}
\label{sec:furtherVisualization}
\begin{figure}[H]
    \centering
    \includegraphics[width=0.98\linewidth]{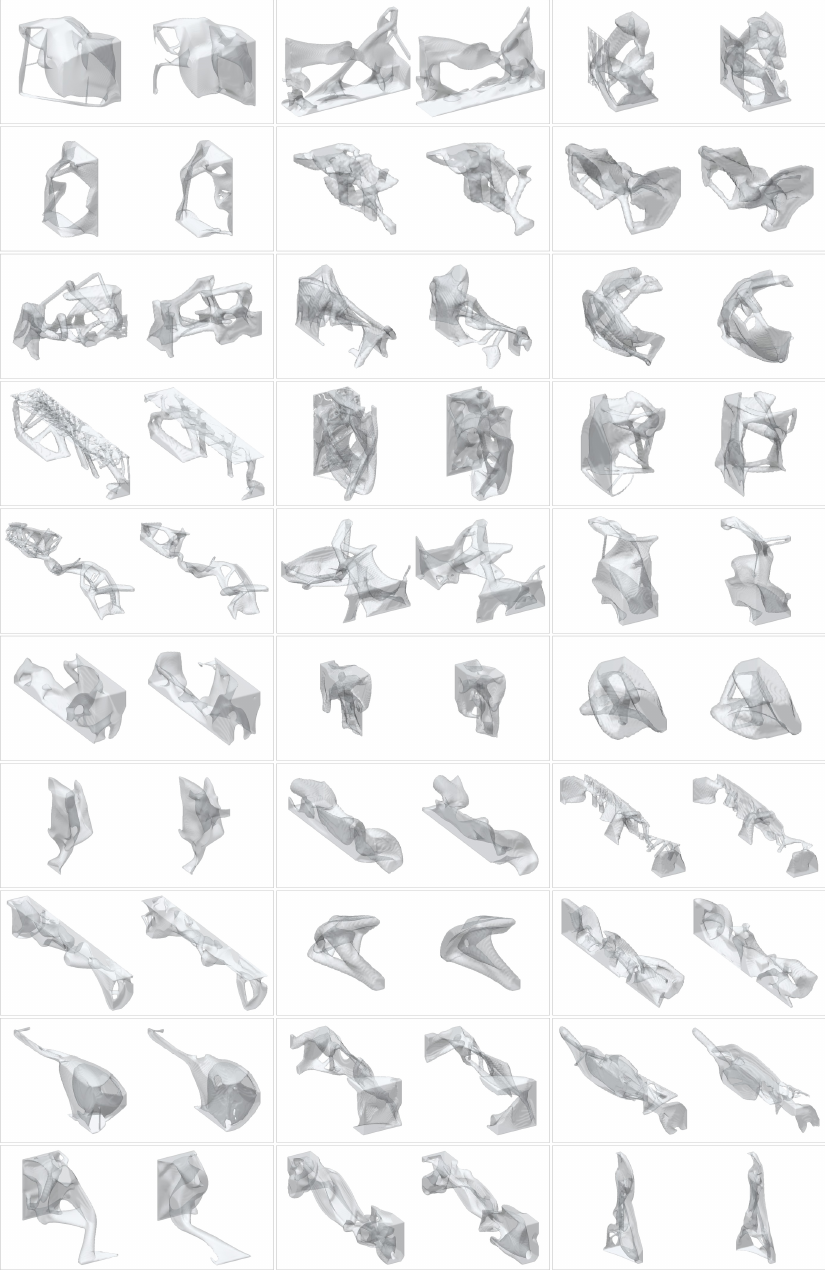}
    \caption{Selection of optimized topologies from the topology optimization (left) and OAT3D (right)}
    \label{fig:contact_vlm_3x10_a}
\end{figure}

\begin{figure}[H]
    \centering
    \includegraphics[width=0.98\linewidth]{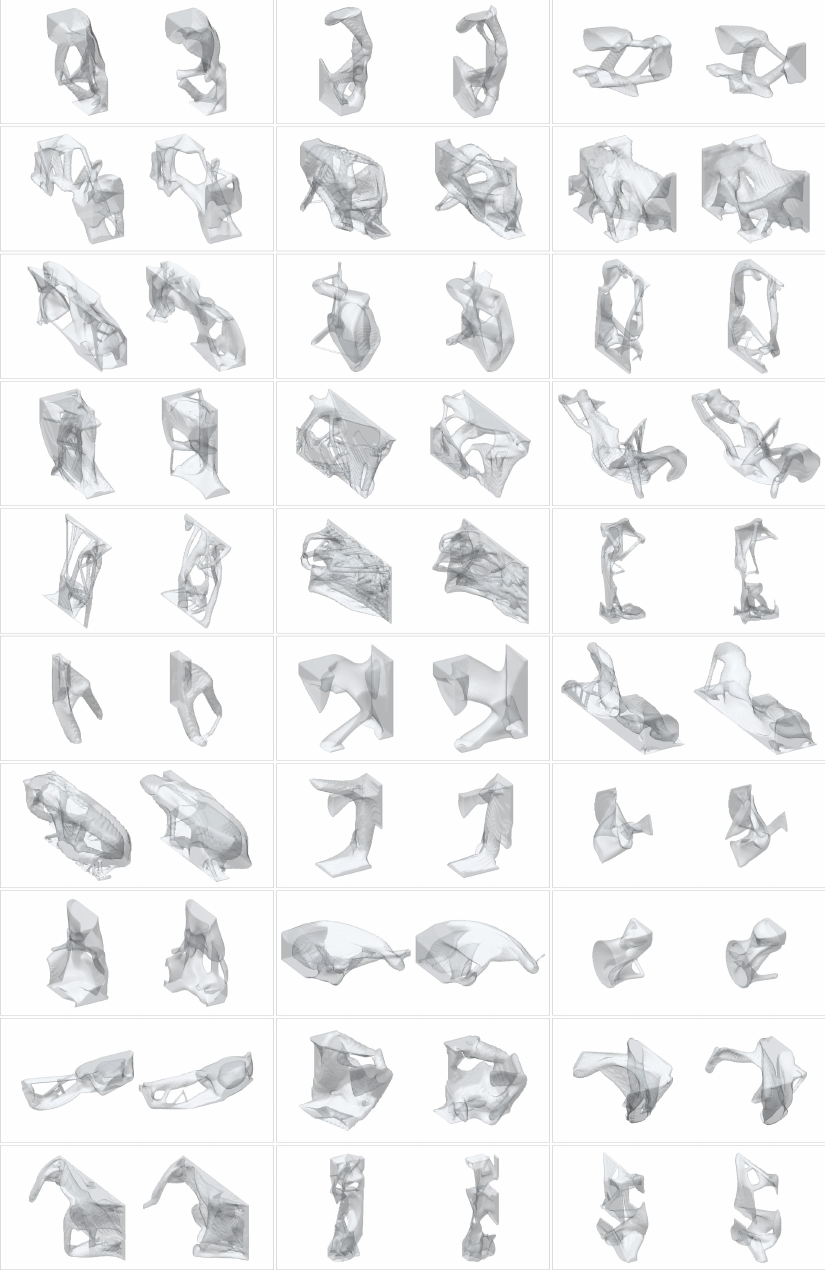}
    \caption{(continued) Selection of optimized topologies from the topology optimization (left) and OAT3D (right)}
    \label{fig:contact_vlm_3x10_b}
\end{figure}

\begin{figure}[H]
    \centering
    \includegraphics[width=0.95\linewidth]{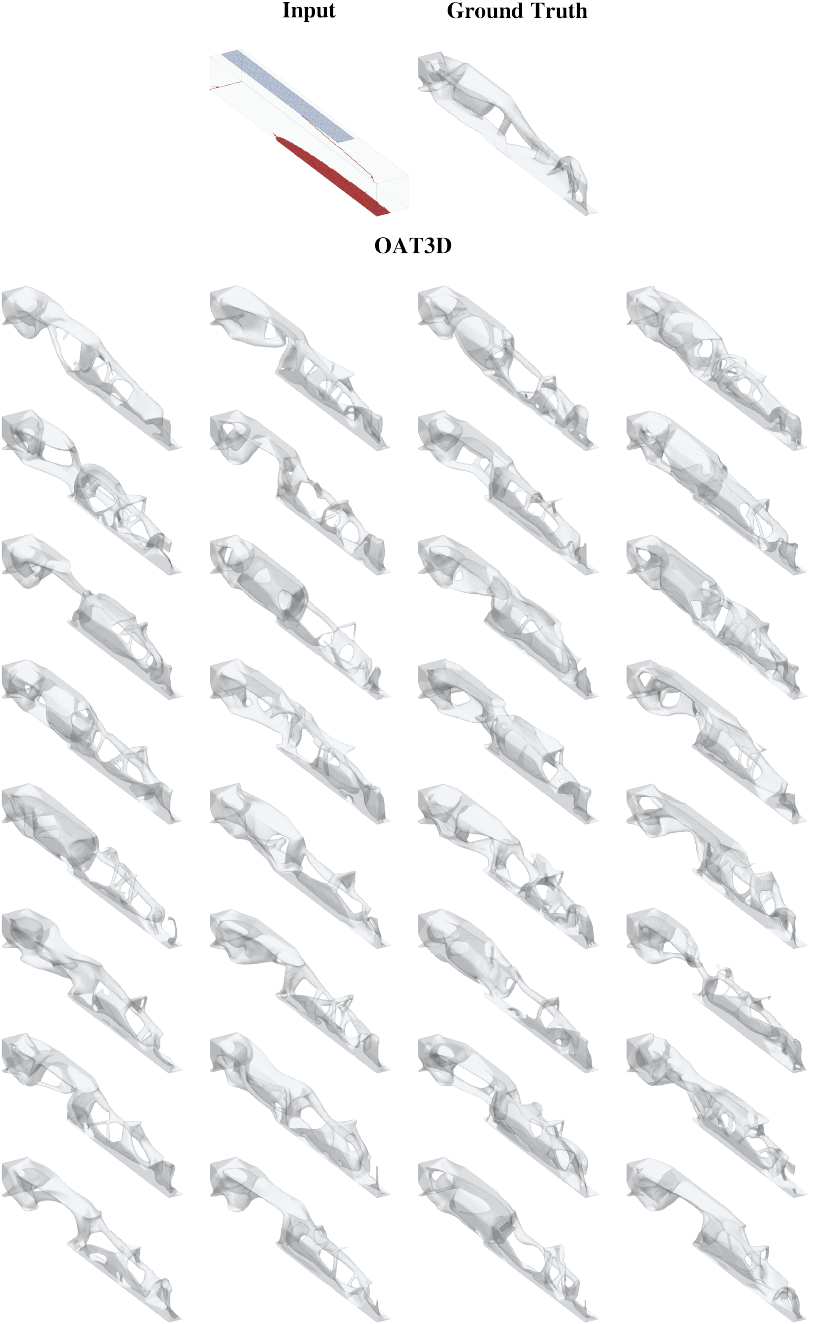}
    \caption{Visualizing all 32 candidates generated by OAT3D given a selected problem. The input boundary condition and the ground truth topology are shown at the top. }
    \label{fig:bestOf32}
\end{figure}

\newpage
\section{Volume Fraction Stratification Study}
\label{sec:volumeStratification}

\paragraph{2D Topology Optimization.} We examine how topology-prediction performance varies with the prescribed volume fraction. All results use single generation per problem, without candidate selection or refinement. We first evaluate the 2D OAT model of \citet{heyrani2026optimize} on 7,500 problems balanced across five volume-fraction bins (Table~\ref{tab:volume-fraction-study}).

\paragraph{3D Topology Optimization.} We repeat this analysis on the 2,000-problem OAT3D evaluation set, stratifying it into three volume-fraction bins and comparing OAT3D with the three NITO-3D baselines (Table~\ref{tab:raw-volume-stratified}).

\paragraph{Findings.} Lower prescribed volume fractions are associated with higher failure rates in both studies. From the highest to the lowest volume-fraction bin, OAT's overall failure rate increases from 32.40\% to 74.00\%, while OAT3D's increases from 24.80\% to 72.12\%. Their median CE among passing predictions also increases, from 2.98\% to 10.06\% and from 19.87\% to 35.33\%, respectively. OAT3D achieves the lowest failure rate in every 3D bin, but low-volume prediction remains challenging: both OAT3D-trained NITO-3D baselines exceed 96\% failure in the lowest bin.

\begin{table}[H]
\centering\small
\caption{Single-generation 2D OAT performance on 7,500 distinct problems, with 1,500 per prescribed-volume-fraction bin. The set combines official-test problems with additional problems generated using the official sampler; the second column gives the number of newly generated problems. Overall failure follows Section~\ref{sec:metrics}: numerical failure, structural failure ($C>2C^{\mathrm{ref}}$), or volume excess ($\widehat v-v^\star>0.10$). Overlapping failures are counted once and failure rates use all problems in each bin. CE statistics exclude failed predictions, and mean VFE is signed.}
\label{tab:volume-fraction-study}
\setlength{\tabcolsep}{5pt}
\begin{tabular}{lrrrrr}
\toprule
\shortstack{Prescribed\\VF}
& \shortstack{Newly\\generated}
& \shortstack{CE mean\\(\%) $\downarrow$}
& \shortstack{CE median\\(\%) $\downarrow$}
& \shortstack{Mean VFE\\(\%)}
& \shortstack{Overall failure\\(\%) $\downarrow$} \\
\midrule
$[0.20,0.30)$ & 1,390 & 16.53 & 10.06 & -0.20 & 74.00 \\
$[0.30,0.40)$ & 510 & 13.01 & 6.86 & -0.14 & 53.07 \\
$[0.40,0.50)$ & 318 & 9.55 & 4.67 & -0.20 & 46.73 \\
$[0.50,0.60)$ & 201 & 7.11 & 3.79 & -0.19 & 40.20 \\
$[0.60,0.70]$ & 81 & 7.14 & 2.98 & -0.06 & 32.40 \\
\bottomrule
\end{tabular}
\end{table}

\begin{table}[H]
\centering\scriptsize
\caption{Single-generation performance on the 2,000-problem OAT3D evaluation set, stratified by prescribed volume fraction. All models are evaluated on the same problems within each bin. The NITO-3D baselines are trained on either the original NITO-3D dataset or OAT3D, with the latter evaluated at 23M and 811M parameters. Failure criteria and metric aggregation follow Table~\ref{tab:volume-fraction-study}; failure rates use all problems in each bin. Dashes indicate that no predictions pass the failure criteria, leaving CE undefined.}
\label{tab:raw-volume-stratified}
\setlength{\tabcolsep}{5pt}
\begin{tabular}{lrrrr}
\toprule
Method
& \shortstack{CE mean\\(\%) $\downarrow$}
& \shortstack{CE median\\(\%) $\downarrow$}
& \shortstack{Mean VFE\\(\%)}
& \shortstack{Overall failure\\(\%) $\downarrow$} \\
\midrule
\multicolumn{5}{l}{\textit{(a) Prescribed VF $[0.05,0.10)$: 312 problems}} \\
NITO (NITO-trained) & - & - & - & 100.00 \\
NITO (OAT-trained, 23M) & 41.84 & 29.67 & 5.20 & 97.44 \\
NITO (OAT-trained, 811M) & 56.64 & 65.13 & $-$4.51 & 96.47 \\
OAT3D & 38.88 & 35.33 & $-$4.29 & 72.12 \\
\midrule
\multicolumn{5}{l}{\textit{(b) Prescribed VF $[0.10,0.20)$: 793 problems}} \\
NITO (NITO-trained) & 56.23 & 56.79 & 39.22 & 99.37 \\
NITO (OAT-trained, 23M) & 44.92 & 40.04 & 5.00 & 93.82 \\
NITO (OAT-trained, 811M) & 45.62 & 33.42 & 3.92 & 91.55 \\
OAT3D & 37.60 & 31.76 & $-$4.71 & 50.57 \\
\midrule
\multicolumn{5}{l}{\textit{(c) Prescribed VF $[0.20,0.30]$: 895 problems}} \\
NITO (NITO-trained) & 30.75 & 14.57 & 15.62 & 99.44 \\
NITO (OAT-trained, 23M) & 41.96 & 41.15 & 3.33 & 81.01 \\
NITO (OAT-trained, 811M) & 43.48 & 42.71 & 0.08 & 79.66 \\
OAT3D & 27.80 & 19.87 & $-$4.19 & 24.80 \\
\bottomrule
\end{tabular}
\end{table}

\newpage
\section{Failure analysis beyond the volume-fraction breakdown}
We further analyze every single-generation prediction on the 2,000 problem evaluation set in three ways: by the failure criterion that is violated, by the structural cause of compliance failures, and by problem attribute.

\begin{enumerate}
    \item \textbf{Failure criterion.} We break down the failures into the three failure modes: failed linear solve, compliance failure ($C > 2 C_\text{ref}$), and volume excess (predicted volume fraction more than 0.10 above the prescribed one). A prediction that violates both the compliance and the volume criterion is counted as a compliance failure (Table~\ref{tab:failure-causes}).
    \item \textbf{Structural cause.} We threshold each prediction and check
    every loaded component independently. A component is restrained if the load applied to it does not act in any direction that is unsupported by the displacement boundary condition. We then measure the share of the applied force that acts on restrained components. In Table~\ref{tab:failure-causes}, generated topologies are assigned the following classes:
        \begin{enumerate}
            \item Incomplete load path: less than 1\% of the force acts on restrained material.
            \item Partial load path: 1\% to 90\% of the force acts on restrained material.
            \item Intact load path: at least 90\%. The design carries the load to the supports and is still more than twice as compliant as the reference.
        \end{enumerate}
    \item \textbf{Problem attributes.} Beyond volume fraction, we break down the test set by number of loaded nodes, loaded regions, fixed nodes, and elements (Table~\ref{tab:failure-attributes}). Each entry is the overall failure rate within a group of problems (the percentage of the problems in that row for which the method's single prediction fails). It is not a share of all failures.
\end{enumerate}

\begin{table}[h]
\centering
\caption{Failed single-generation predictions on the OAT test set by structural cause, as a percentage of the problems in each volume-fraction range. The five cause columns are exclusive and sum to the overall failure rate. Numbers in parentheses are shares of compliance failures in each class.}
\label{tab:failure-causes}
\scriptsize
\resizebox{\textwidth}{!}{{
\begin{tabular}{lrlrrrrrr}
\toprule
 & & & \multicolumn{6}{c}{Failure (\%) $\downarrow$} \\
  \cmidrule(lr){4-9}
 & & & & & \multicolumn{3}{c}{Compliance failure} & \\
  \cmidrule(lr){6-8}
 & & & & Solve & Incomplete & Partial & Intact & Volume \\
Volume fraction & Problems & Method & Overall & failed & load path & load path & load path & failure only \\
\midrule
$0.05 \le v^* < 0.10$ & 312 & NITO-3D (NITO-trained) & 100 & 1.92 & 43.3 & 48.4 & 5.77 & 0.64 \\
 &  &  &  &  & (44.4) & (49.7) & (5.9) &  \\
 &  & NITO-3D (OAT-trained, 23M) & 97.4 & 0.00 & 46.8 & 36.2 & 14.4 & 0.00 \\
 &  &  &  &  & (48.0) & (37.2) & (14.8) &  \\
 &  & NITO-3D (OAT-trained, 811M) & 96.5 & 0.00 & 40.7 & 40.1 & 15.7 & 0.00 \\
 &  &  &  &  & (42.2) & (41.5) & (16.3) &  \\
 &  & OAT3D & 71.8 & 0.00 & 0.96 & 6.41 & 64.4 & 0.00 \\
 &  &  &  &  & (1.4) & (8.9) & (89.7) &  \\
\midrule
$0.10 \le v^* < 0.20$ & 793 & NITO-3D (NITO-trained) & 99.4 & 2.40 & 41.4 & 50.7 & 4.67 & 0.25 \\
 &  &  &  &  & (42.8) & (52.4) & (4.8) &  \\
 &  & NITO-3D (OAT-trained, 23M) & 93.8 & 0.00 & 17.2 & 44.4 & 32.3 & 0.00 \\
 &  &  &  &  & (18.3) & (47.3) & (34.4) &  \\
 &  & NITO-3D (OAT-trained, 811M) & 91.6 & 0.00 & 13.4 & 38.7 & 39.5 & 0.00 \\
 &  &  &  &  & (14.6) & (42.3) & (43.1) &  \\
 &  & OAT3D & 50.3 & 0.00 & 0.50 & 2.14 & 47.5 & 0.13 \\
 &  &  &  &  & (1.0) & (4.3) & (94.7) &  \\
\midrule
$0.20 \le v^* \le 0.30$ & 895 & NITO-3D (NITO-trained) & 99.4 & 2.46 & 21.5 & 65.1 & 9.27 & 1.12 \\
 &  &  &  &  & (22.4) & (67.9) & (9.7) &  \\
 &  & NITO-3D (OAT-trained, 23M) & 81.0 & 0.00 & 3.58 & 27.9 & 49.5 & 0.00 \\
 &  &  &  &  & (4.4) & (34.5) & (61.1) &  \\
 &  & NITO-3D (OAT-trained, 811M) & 79.7 & 0.00 & 3.46 & 21.0 & 55.1 & 0.11 \\
 &  &  &  &  & (4.4) & (26.4) & (69.2) &  \\
 &  & OAT3D & 24.8 & 0.00 & 0.22 & 1.34 & 23.2 & 0.00 \\
 &  &  &  &  & (0.9) & (5.4) & (93.7) &  \\
\bottomrule
\end{tabular}}
}
\end{table}

\begin{table}[h]
\centering
\caption{Overall single-generation failure rate by problem attribute on OAT test set. Each entry is the percentage of the problems in that row that fail.}
\label{tab:failure-attributes}
\scriptsize
{
\begin{tabular}{llrrrrr}
\toprule
 & & & \multicolumn{4}{c}{Failure (\%) $\downarrow$} \\
  \cmidrule(lr){4-7}
 & & & NITO-3D & NITO-3D & NITO-3D & \\
Attribute & Range & Problems & (NITO-trained) & (OAT, 23M) & (OAT, 811M) & OAT3D \\
\midrule
Loaded nodes & 1 & 125 & 95.2 & 56.8 & 52.8 & 12.8 \\
 & 2--100 & 336 & 99.1 & 75.6 & 72.6 & 24.1 \\
 & 101--1,000 & 672 & 99.9 & 91.2 & 90.8 & 43.6 \\
 & 1,001--10,000 & 677 & 100 & 96.5 & 95.0 & 55.4 \\
 & $>$ 10,000 & 190 & 100 & 95.8 & 93.2 & 43.7 \\
\midrule
Load regions & 1 & 369 & 97.6 & 73.4 & 70.7 & 21.4 \\
 & 2 & 371 & 99.7 & 81.7 & 78.2 & 30.5 \\
 & 3 & 329 & 100 & 91.5 & 90.3 & 40.4 \\
 & 4 & 258 & 100 & 92.6 & 92.6 & 48.8 \\
 & 5--6 & 373 & 100 & 98.1 & 96.8 & 54.4 \\
 & $\ge$ 7 & 300 & 100 & 97.7 & 97.3 & 64.7 \\
\midrule
Fixed nodes & $\le$ 100 & 68 & 100 & 94.1 & 91.2 & 33.8 \\
 & 101--1,000 & 405 & 99.3 & 93.6 & 91.4 & 49.9 \\
 & 1,001--10,000 & 1,028 & 99.8 & 90.3 & 89.3 & 44.7 \\
 & $>$ 10,000 & 499 & 99.0 & 80.6 & 78.2 & 32.7 \\
\midrule
Elements & $<$ 150k & 185 & 100 & 89.2 & 86.5 & 34.6 \\
 & 150k--250k & 242 & 100 & 88.8 & 85.5 & 42.1 \\
 & 250k--400k & 371 & 99.5 & 87.9 & 86.3 & 42.6 \\
 & 400k--600k & 480 & 99.6 & 89.2 & 88.1 & 44.6 \\
 & 600k--800k & 379 & 98.4 & 87.6 & 86.8 & 42.5 \\
 & $>$ 800k & 343 & 100 & 89.5 & 87.8 & 43.4 \\
\bottomrule
\end{tabular}
}
\end{table}

\begin{figure}[H]
    \centering
    \includegraphics[width=0.9\linewidth]{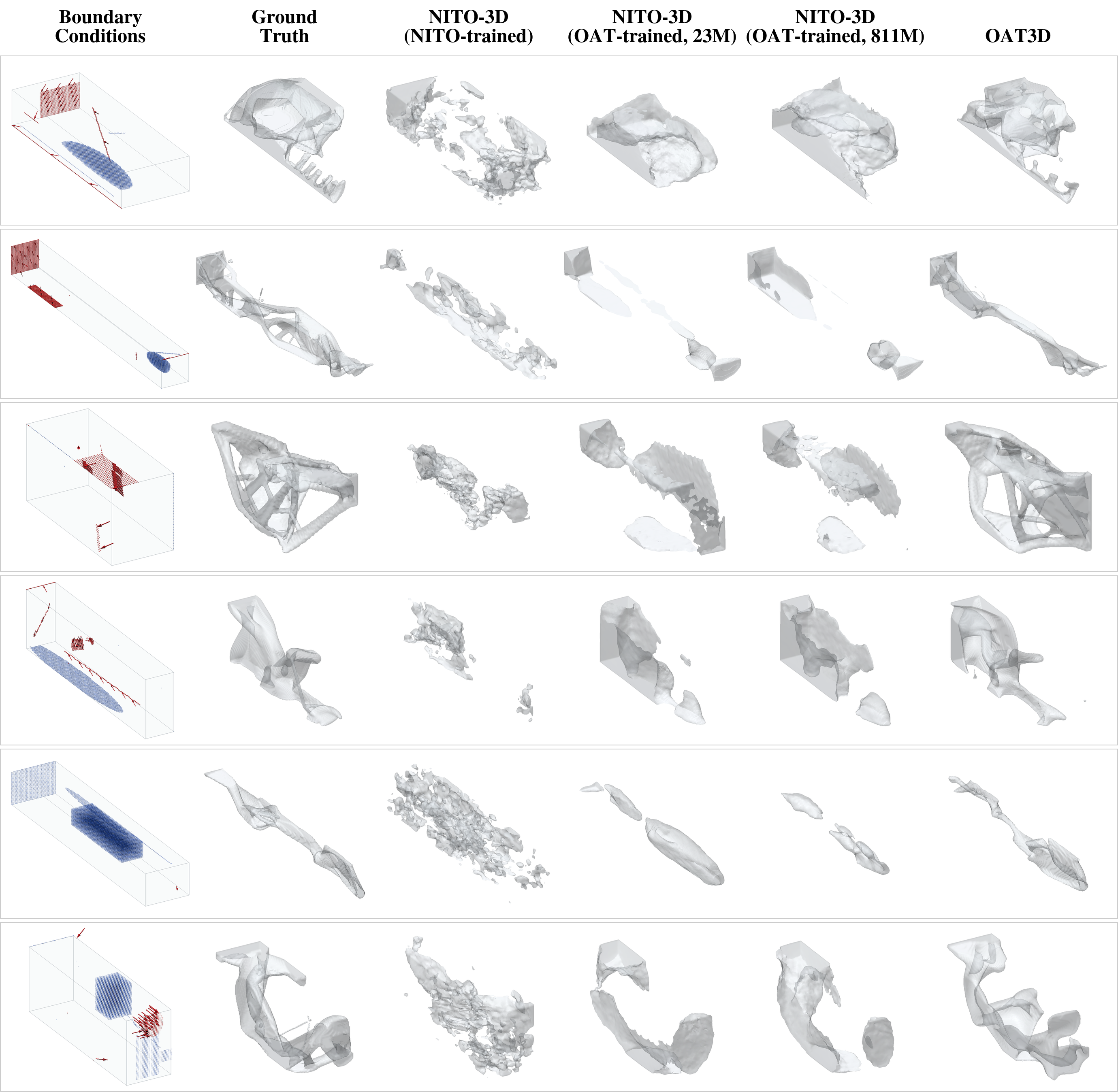}
    \caption{Single-generation predictions on sample test problems across all models.}
    \label{fig:nitoFailure}
\end{figure}

\paragraph{Findings.} Almost all failures are compliance failures (Table~\ref{tab:failure-causes}). Failed linear solves occur only for NITO-3D trained on its own dataset (2.35\% of problems), and volume-excess-alone accounts for at most 0.70\% of problems for any method. The models differ in why the compliance criterion is violated. Across all volume fractions, OAT3D demonstrates superior ability in generating topologies with intact load paths compared to the other baselines. The NITO-3D baselines, in contrast, often do not form a complete load path at all, visually demonstrated in Figure~\ref{fig:nitoFailure}.

Across problem attributes (Table~\ref{tab:failure-attributes}), failure increases with the number of loaded nodes and number of loaded regions. For OAT3D it rises from 12.8\% for a single loaded node to 55.4\% for 1,001-10,000 loaded nodes, and from 21.4\% for one load region to 64.7\% for seven or more. Above 100 fixed nodes, failure decreases as the number of fixed nodes grows for OAT3D. The number of elements shows no clear trend for any model.

\newpage
\section{Post Optimizer and Initialization Ablation}
We perform post-optimization on the same initialization with Projected Gradient Descent (PGD) and Optimality Criteria (OC). Table~\ref{tab:PO-ablation}  outlines the structural compliance across post-optimization steps. PGD is a worse post-optimizer due to how it scales its updates. PGD normalizes its step globally, by the largest sensitivity magnitude and then by spectral estimates that take several iterations to calibrate. This is acceptable over a full optimization run, but not within a budget of a few updates. This matters because OAT3D outputs are strongly quantized, near-binary fields. At full penalization, near-void elements have almost no sensitivity. We therefore ramp the penalty factor from 1 to 3 over the first five iterations, starting from the raw-inference design that is blended with a 25\% uniform field at the prescribed volume fraction. This lifts every element off zero while keeping the predicted topology dominant.
\begin{table}[H]
\centering
\caption{Comparison of Optimality Criteria (OC) and Projected Gradient Descent (PGD) as post optimizers following best-of-2 selection.}
\label{tab:PO-ablation}
\scriptsize
\setlength{\tabcolsep}{3pt}
{
\begin{tabular}{lccrrrr}
\toprule
Method & N & PO Steps & CE mean (\%) $\downarrow$ & CE median (\%) $\downarrow$ & Mean VFE (\%) & Failure (\%) $\downarrow$ \\
\midrule
\multicolumn{7}{l}{\textit{OAT-initialized}}\\
OAT3D + OC  & 2 & 5 & 17.01 & 12.30 & 0.00 & 18.35 \\
OAT3D + OC  & 2 & 10 & 2.71 & 1.42 & 0.00 & 6.45 \\
OAT3D + OC  & 2 & 15 & 1.15 & 0.50 & 0.00 & 3.35 \\
\midrule
OAT3D + PGD & 2 & 5 & 59.18 & 48.54 & 0.00 & 41.50 \\
OAT3D + PGD & 2 & 10 & 18.00 & 14.54 & 0.00 & 5.30 \\
OAT3D + PGD & 2 & 15 & 11.04 & 7.83 & 0.00 & 2.75 \\
\bottomrule
\end{tabular}
}
\end{table}

\newpage
\section{Study of Timing and Iteration Count}

We assess computational scaling across all 2,000 test cases, grouped by element count and boundary-condition percentage. Tables~\ref{tab:dense_time_by_elements_bc}-\ref{tab:sparse_time_by_elements_bc} show relatively stable generation times across problem size and percent occupied by boundary conditions, while PGD (Table~\ref{tab:pgd_steps_by_elements_bc}) cost per step increases with problem size. Table~\ref{tab:speedup_raw_pgd} reports mean per-case speedups for raw generation. For raw generation, the speedup is most prominent in higher element counts. Table~\ref{tab:additional_speedup_with_PO} accounts for best-of-2 generation including both compliance evaluations and 10 OC post-optimization steps. Once post-optimization is added, the speedup is more pronounced in higher element counts.  

\begin{table}[H]
\centering
{
\caption{Mean dense-stage time in seconds per generated topology.}
\label{tab:dense_time_by_elements_bc}
\normalsize
\begin{tabular}{lrrrrr}
\toprule
 & \multicolumn{5}{c}{Boundary-condition percentage (BC, \%)} \\
\cmidrule(lr){2-6}
Elements (k) & $<0.5$ & $[0.5,1)$ & $[1,2)$ & $[2,3.5)$ & $\geq 3.5$ \\
\midrule
$<250$      & 0.605 & 0.612 & 0.615 & 0.615 & 0.619 \\
$[250,400)$ & 0.621 & 0.617 & 0.620 & 0.628 & 0.628 \\
$[400,600)$ & 0.621 & 0.619 & 0.627 & 0.631 & 0.636 \\
$[600,800)$ & 0.622 & 0.623 & 0.626 & 0.639 & 0.644 \\
$\geq 800$  & 0.619 & 0.623 & 0.634 & 0.643 & 0.653 \\
\bottomrule
\end{tabular}
}
\end{table}

\begin{table}[H]
\centering
{
\caption{Mean sparse-stage time in seconds per generated topology.}
\label{tab:sparse_time_by_elements_bc}
\normalsize
\begin{tabular}{lrrrrr}
\toprule
 & \multicolumn{5}{c}{Boundary-condition percentage (BC, \%)} \\
\cmidrule(lr){2-6}
Elements (k) & $<0.5$ & $[0.5,1)$ & $[1,2)$ & $[2,3.5)$ & $\geq 3.5$ \\
\midrule
$<250$      & 0.570 & 0.554 & 0.552 & 0.555 & 0.553 \\
$[250,400)$ & 0.544 & 0.562 & 0.561 & 0.552 & 0.566 \\
$[400,600)$ & 0.567 & 0.575 & 0.568 & 0.572 & 0.579 \\
$[600,800)$ & 0.586 & 0.595 & 0.602 & 0.603 & 0.611 \\
$\geq 800$  & 0.622 & 0.621 & 0.605 & 0.616 & 0.637 \\
\bottomrule
\end{tabular}
}
\end{table}

\begin{table}[H]
\centering
{
\caption{Mean PGD iterations per run for topology optimization of the test set on H200 using the dataset-generation pipeline. Mean time per iteration (s) in parentheses.}
\label{tab:pgd_steps_by_elements_bc}
\normalsize
\setlength{\tabcolsep}{1.5pt}
\begin{tabular}{@{}lrrrrr@{}}
\toprule
 & \multicolumn{5}{c}{Boundary-condition percentage (BC, \%)} \\
\cmidrule(lr){2-6}
Elements (k) & $<0.5$ & $[0.5,1)$ & $[1,2)$ & $[2,3.5)$ & $\geq 3.5$ \\
\midrule
$<250$      &  94.6 (0.207) & 106.3 (0.196) & 107.7 (0.183) &  99.4 (0.169) & 113.8 (0.153) \\
$[250,400)$ & 105.6 (0.345) & 123.5 (0.339) & 123.9 (0.309) & 117.3 (0.277) & 103.9 (0.278) \\
$[400,600)$ & 149.3 (0.477) & 129.3 (0.462) & 130.7 (0.421) & 124.6 (0.410) & 131.8 (0.393) \\
$[600,800)$ & 140.2 (0.631) & 131.2 (0.587) & 122.7 (0.590) & 115.0 (0.519) & 129.1 (0.490) \\
$\geq 800$  & 125.9 (0.809) & 151.5 (0.788) & 153.5 (0.745) & 125.5 (0.722) & 140.4 (0.650) \\
\bottomrule
\end{tabular}
}
\end{table}

\begin{table}[H]
\centering
{
\caption{Mean per-case speedup of OAT3D raw generation of the 2,000-problem test set compared to topology optimization on H200s.}
\label{tab:speedup_raw_pgd}
\normalsize
\begin{tabular}{lrrrrr}
\toprule
 & \multicolumn{5}{c}{Boundary-condition percentage (BC, \%)} \\
\cmidrule(lr){2-6}
Elements (k) & $<0.5$ & $[0.5,1)$ & $[1,2)$ & $[2,3.5)$ & $\geq 3.5$ \\
\midrule
$<250$      & $16.5\times$ & $17.7\times$ & $17.6\times$ & $13.8\times$ & $14.3\times$ \\
$[250,400)$ & $31.3\times$ & $35.6\times$ & $31.4\times$ & $26.1\times$ & $23.3\times$ \\
$[400,600)$ & $58.8\times$ & $48.8\times$ & $43.0\times$ & $42.3\times$ & $39.9\times$ \\
$[600,800)$ & $71.0\times$ & $63.1\times$ & $56.9\times$ & $47.6\times$ & $48.4\times$ \\
$\geq 800$  & $80.2\times$ & $93.2\times$ & $90.9\times$ & $67.0\times$ & $63.7\times$ \\
\bottomrule
\end{tabular}
}
\end{table}

\begin{table}[H]
\centering
{
\caption{Speedup over the test set topology optimization vs best-of-2 generation, both candidate compliance evaluations, and 10 OC post-optimization steps on H200s.}
\label{tab:additional_speedup_with_PO}
\normalsize
\begin{tabular}{lrrrrr}
\toprule
 & \multicolumn{5}{c}{Boundary-condition percentage (BC, \%)} \\
\cmidrule(lr){2-6}
Elements (k) & $<0.5$ & $[0.5,1)$ & $[1,2)$ & $[2,3.5)$ & $\geq 3.5$ \\
\midrule
$<250$      & $2.85\times$ & $3.10\times$ & $2.90\times$ & $2.62\times$ & $2.76\times$ \\
$[250,400)$ & $3.27\times$ & $3.72\times$ & $3.76\times$ & $3.57\times$ & $3.18\times$ \\
$[400,600)$ & $3.95\times$ & $3.69\times$ & $3.57\times$ & $3.85\times$ & $4.23\times$ \\
$[600,800)$ & $3.82\times$ & $3.37\times$ & $3.77\times$ & $3.61\times$ & $4.18\times$ \\
$\geq 800$  & $3.05\times$ & $4.05\times$ & $4.36\times$ & $4.03\times$ & $4.57\times$ \\
\bottomrule
\end{tabular}
}
\end{table}

\newpage
\section{Out-of-Distribution Evaluation}

\subsection{Practical Engineering Boundary Conditions}

Some of the common topology optimization problems include the MBB beam, middle and bottom loaded cantilever beam, and bridge. We configure these common topology optimization problems with varying resolutions and run them at 0.3 volume fraction. On average, the cases achieved a compliance error of 23.84\% and a volume fraction error of 3.77\%.

\begin{figure}[H]
    \centering
    \includegraphics[width=0.95\linewidth]{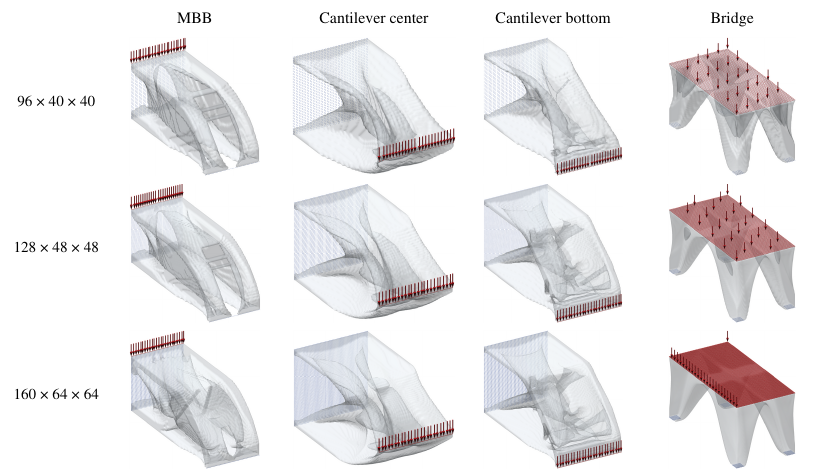}
    \caption{The MBB beam, cantilever beam with middle and bottom load and bridge with conventional topology optimization and PGD as the optimizer. We record the compliance and volume fraction under each rendering.}
    \label{fig:mbb_pgd}
\end{figure}

\begin{figure}[H]
    \centering
    \includegraphics[width=0.95\linewidth]{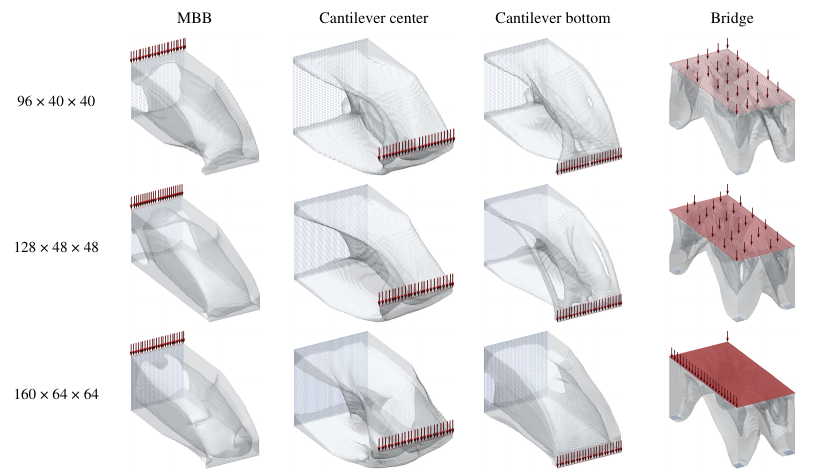}
    \caption{The MBB beam, cantilever beam with middle and bottom load and bridge with OAT3D and no post-optimization. We evaluate the compliance and volume fraction under each rendering.}
    \label{fig:mbb_oat3d}
\end{figure}

\newpage
\subsection{NITO-3D Test Set}
We evaluated OAT3D results on 2,000 samples from the NITO-3D test set, and we report it below together with the one adaptation we found helpful. We did this evaluation with the NITO dataset completely unseen (no training or fine-tuning). We use the same evaluation standards as Section~\ref{sec:evaluation}. We report single generation and best-of-8 as in Section~\ref{sec:best-of-N}.

There are three factors that contribute to NITO-3D dataset being outside of OAT3D's training distribution:
\begin{enumerate}
  \item{Domain size:} NITO-3D uses seven fixed grids of 32,000 to 48,000 elements. OAT3D is trained on domains of at least 65,536 elements whose sides are at least 32 elements long. Every NITO-3D grid is below the minimum element count. Six of the seven have a side shorter than 32 elements.
  \item{Volume fraction:} NITO-3D prescribes $0.15$ to $0.75$, and 84\% of the problems in this evaluation are above OAT3D dataset's maximum volume fraction of $0.30$.
  \item{Boundary condition:} NITO-3D has its own boundary condition sampler.
\end{enumerate}

Without any adaptation, OAT3D fails on 40.35\% of the 2,000 problems with a single generation. The dominant failure is excess volume, not a non-functional structure: for $v^* \le 0.30$, 45.4\% of the designs exceed the volume tolerance. We regard this as a limitation: OAT3D does not transfer directly to domain sizes smaller than those it was trained on.

\begin{table}[h]
\centering
\caption{Direct evaluation of OAT3D generation on the NITO-3D test set.}
\label{tab:nito-direct}
\scriptsize
{
\begin{tabular}{llrrrrrrrr}
\toprule
  & & & \multicolumn{2}{c}{CE (\%) $\downarrow$} & Mean VFE & \multicolumn{3}{c}{Failure (\%) $\downarrow$} \\
  \cmidrule(lr){4-5}\cmidrule(lr){7-9}
  Volume fraction & Problems & Selection & Mean & Median & (\%) & Volume & Compliance & Overall \\
\midrule
$v^* \le 0.30$        & 315   & Single    & 14.89 &  8.22 & $+38.89$ & 45.4 &  9.8 & 54.29 \\
                      &       & Best-of-8 &  9.42 &  3.99 & $+51.20$ & 55.2 &  6.0 & 60.63 \\
\midrule
$0.30 < v^* \le 0.50$ & 678   & Single    & 16.97 &  9.99 & $+13.09$ & 38.6 &  9.1 & 47.35 \\
                      &       & Best-of-8 & 14.99 &  6.71 & $+24.46$ & 53.7 &  3.4 & 57.08 \\
\midrule
$v^* > 0.50$          & 1,007 & Single    & 21.26 & 12.96 & $-30.06$ & 11.9 & 19.7 & 31.28 \\
                      &       & Best-of-8 & 18.13 &  8.96 & $-15.15$ & 27.0 &  9.0 & 36.05 \\
\midrule
Full split            & 2,000 & Single    & 19.21 & 11.13 & $-4.57$  & 26.2 & 14.6 & 40.35 \\
                      &       & Best-of-8 & 16.25 &  7.70 & $+8.73$  & 40.5 &  6.7 & 47.05 \\
\bottomrule
\end{tabular}

}
\end{table}

One possible adaptation is to pose each problem on a grid refined by a factor of 2 in every direction. The load and supports stay at the same physical positions, $v^*$ is unchanged, and the model is unchanged. The refined grids have 256,000 to 384,000 elements, which is inside OAT3D model's trained range. To remain on the NITO-3D benchmark itself, each $2{\times}2{\times}2$ block of the generated design is averaged back to one NITO-3D element, and the result is scored on the NITO-3D grid against the same references as Table~\ref{tab:nito-direct}. We can see that volume failure reduces dramatically, even dropping to single digits in the $v^* \le 0.30$ range (Table~\ref{tab:nito-refined}). With this adaptation, OAT3D generation underfills the domain, but across all volume ranges of the NITO-3D test set, the compliance error is actually lower than OAT3D generation evaluated on the OAT3D test set. After 5 updates of post-optimization, the median compliance error is 0.89\% on the entire 2,000-problem test set.

\begin{table}[H]
\centering
\caption{Evaluation of OAT3D generation with $\times 2$ mesh refinement on the NITO-3D test set, averaged back
and scored on the NITO-3D grid.}
\label{tab:nito-refined}
\scriptsize
{
\begin{tabular}{lrlrrrrrr}
\toprule
  & & & \multicolumn{2}{c}{CE (\%) $\downarrow$} & Mean VFE & \multicolumn{3}{c}{Failure (\%) $\downarrow$} \\
  \cmidrule(lr){4-5}\cmidrule(lr){7-9}
  Volume fraction & Problems & Selection & Mean & Median & (\%) & Volume & Compliance & Overall \\
\midrule
$v^* \le 0.30$        & 315   & Single        &  18.3 &  11.4 & $-10.8$  & 0.6 &  6.7 &  7.30 \\
                      &       & Best-of-8     &  12.1 &  7.00 & $-7.12$  & 0.3 &  4.4 &  4.76 \\
\midrule
$0.30 < v^* \le 0.50$ & 678   & Single        &  23.3 &  15.1 & $-31.8$  & 0.0 & 16.7 &  16.7 \\
                      &       & Best-of-8     &  18.4 &  10.9 & $-28.2$  & 0.6 &  9.7 &  10.3 \\
\midrule
$v^* > 0.50$          & 1,007 & Single        &  38.1 &  32.0 & $-71.8$  & 0.0 & 39.4 &  39.4 \\
                      &       & Best-of-8     &  34.7 &  27.7 & $-69.0$  & 0.0 & 25.3 &  25.3 \\
\midrule
Full split            & 2,000 & Single        &  28.5 &  19.7 & $-48.7$  & 0.1 & 26.6 &  26.7 \\
                      &       & Best-of-8     &  24.6 &  15.0 & $-45.4$  & 0.2 & 16.8 &  17.0 \\
                      &       & Single + 5 OC &  1.80 &  0.89 & $0.00$   & 0.0 &  0.0 &  0.00 \\
\bottomrule
\end{tabular}
}
\end{table}

\clearpage
\section{Implementation of OAT3D Applications}
\label{sec:oat3d_applications_appendix}

\subsection{Joint-count control}
\label{sec:oat3d_joint_control_appendix}

\paragraph{Joint definition.}
We use branch junctions in the structural skeleton as a measure of structural complexity. Reference densities are thresholded at 0.3 and skeletonized with the C++ VoxelCore implementation. Candidate joints are curve-skeleton vertices of degree at least three; medial-sheet edges are excluded. Nearby candidates are merged using the interior-distance radius $r_i$ at each candidate. For candidates $i$ and $j$, the graph distance must satisfy
\begin{equation}
    d_G(i,j) \leq \min\!\left(8,\,2\min(r_i,r_j)\right),
    \label{eq:oat3d_joint_merge}
\end{equation}
with distances measured in native voxel units. Every pair in a merged group must satisfy this bound. A connecting path must have radius at least $0.5\min(r_i,r_j)$, sampled at intervals no larger than one voxel, and cannot pass through an unrelated candidate. Each merged group is counted as one joint. Generated structures are thresholded at 0.3 and restricted to their largest six-connected solid component before skeletonization and counting.

\paragraph{Joint-count regressors.}
Following TopoCtrl~\citep{chen2026topoctrl}, we train separate 3D CNNs to predict joint count from the clean dense and sparse latents. The dense input has 8 channels on a $16^3$ grid. Sparse 16-channel features are scattered onto a $64^3$ grid with an additional active-cell mask. Each network uses channel widths 16, 32, 64, and 128, three residual downsampling blocks with GroupNorm and SiLU, and global mean and max pooling followed by a scalar MLP.  We predict the standardized log count
\begin{equation}
    q(J)=\frac{\log(1+J)-\mu_J}{s_J},
    \qquad \mu_J=2.511887,\quad s_J=1.775882,
    \label{eq:oat3d_joint_target}
\end{equation}
where $\mu_J$ and $s_J$ are computed from the training labels. Training minimizes squared prediction error on 496,858 labeled training structures, using one of eight reflected views per structure per epoch. We use AdamW with learning rate $3\times10^{-4}$, weight decay $10^{-4}$, batch size 256, a 200-step warmup, and gradient-norm clipping at 1.0. We use checkpoints saved after 30 and 40 epochs for the dense and sparse regressors, respectively.

\paragraph{Guided generation.}
For each stage, let $z_{\mathrm{ref}}$ be the normalized latent of the original structure and $f$ the trained regressor. At noise level $\sigma$, the estimated clean latent and the two guidance losses are
\begin{align}
    \widehat z_0 &= z_\sigma-\sigma v_{\mathrm{CFG}}(z_\sigma,\sigma), \\
    \mathcal{L}_J &= \bigl(f(\widehat z_0)-q(J^\star)\bigr)^2, \\
    \mathcal{L}_{\mathrm{ref}} &= \operatorname{MSE}(\widehat z_0,z_{\mathrm{ref}}).
    \label{eq:oat3d_joint_guidance_losses}
\end{align}
Before each Euler step, we apply
\begin{equation}
    z'_\sigma=z_\sigma-\nabla_{z_\sigma}\!\left(100\mathcal{L}_J+100\mathcal{L}_{\mathrm{ref}}\right).
    \label{eq:oat3d_joint_guidance_update}
\end{equation}
Gradients pass through the frozen generator and regressor. We recompute the velocity at $z'_\sigma$ before taking the Euler step. Reference MSE averages over all dense latent features; in the sparse stage, it uses only coordinates shared with the reference and is zero if none match. Sparse guidance updates feature values while keeping the active dense occupancy blocks fixed.

Both stages start from full Gaussian noise and retain the original problem conditioning. We use 20 Euler steps with scheduler shift 6, dense CFG 5, and sparse CFG 2.5, without clipping or normalizing the guidance gradients. For each problem, we request $J^\star\in\{5,10,20,50,100\}$ and draw 32 candidates per target, using the same candidate seeds across targets. The joint-control examples are selected from the first 100 eligible held-out problems.

\newpage
\subsection{Lattice replacement}
\label{sec:oat3d_lattice_edit_appendix}

\paragraph{Edited structure.}
We adapt the lattice replacement procedure of TopoEdit~\citep{chen2026topoedit} to OAT3D. Starting from the ground truth density field thresholded at 0.3, we compute the convex hull of the occupied voxel centers and intersect it with a binary body-centered cubic or simple cubic lattice. The cell pitch is 16 native voxels and the member diameter is 4. Body-centered cubic members connect each cell center to its corners; simple cubic members follow the cell edges. Cell corners are aligned to multiples of 16 in native coordinates, and occupancy is evaluated at voxel centers. This construction fills the convex hull without retaining an outer shell. We encode the edited density field with the dense and sparse VAEs. Loads, supports, domain dimensions, and BNG conditioning remain unchanged; the volume-fraction input becomes the mean density of the lattice filled structure.

\paragraph{Guided editing.}
Let $z_{\mathrm{edit}}$ denote a normalized latent of the edited structure. We initialize
\begin{equation}
    z_{\sigma_s}=(1-\sigma_s)z_{\mathrm{edit}}+\sigma_s\epsilon,
    \qquad \epsilon\sim\mathcal{N}(0,I),\quad \sigma_s\approx0.427314,
    \label{eq:oat3d_lattice_initialization}
\end{equation}
and take the final three Euler updates of a 20-step schedule with shift 6 in both stages. Dense and sparse CFG weights are 5 and 2.5. At each update, we compute the clean estimate $\widehat z_0$ as above and apply
\begin{equation}
    z'_\sigma=z_\sigma-100\nabla_{z_\sigma}\operatorname{MSE}(\widehat z_0,z_{\mathrm{edit}}).
    \label{eq:oat3d_lattice_guidance}
\end{equation}
The gradient passes through the generator's clean-latent output. We recompute the velocity after this correction and then take the Euler step. Generator weights remain fixed, and guidance gradients are neither clipped nor normalized.

Decoded dense occupancy above 0.3, cropped to the original domain, determines the sparse active blocks. Edited sparse features are aligned to these blocks by coordinate, with zero reference values at coordinates absent from the edited target. Sparse guidance uses only shared coordinates, and the active dense blocks remain fixed during sparse generation.

We use 16 candidate seeds per lattice type and problem. The examples are drawn from the first 500 eligible held-out problems after the first 2,000 entries, with the same domain-dimension criterion as joint control. Both application figures present visually selected examples. Finite element evaluation was not performed for these application experiments.

\newpage

\begin{figure}[H]
    \centering
    \includegraphics[width=\linewidth]{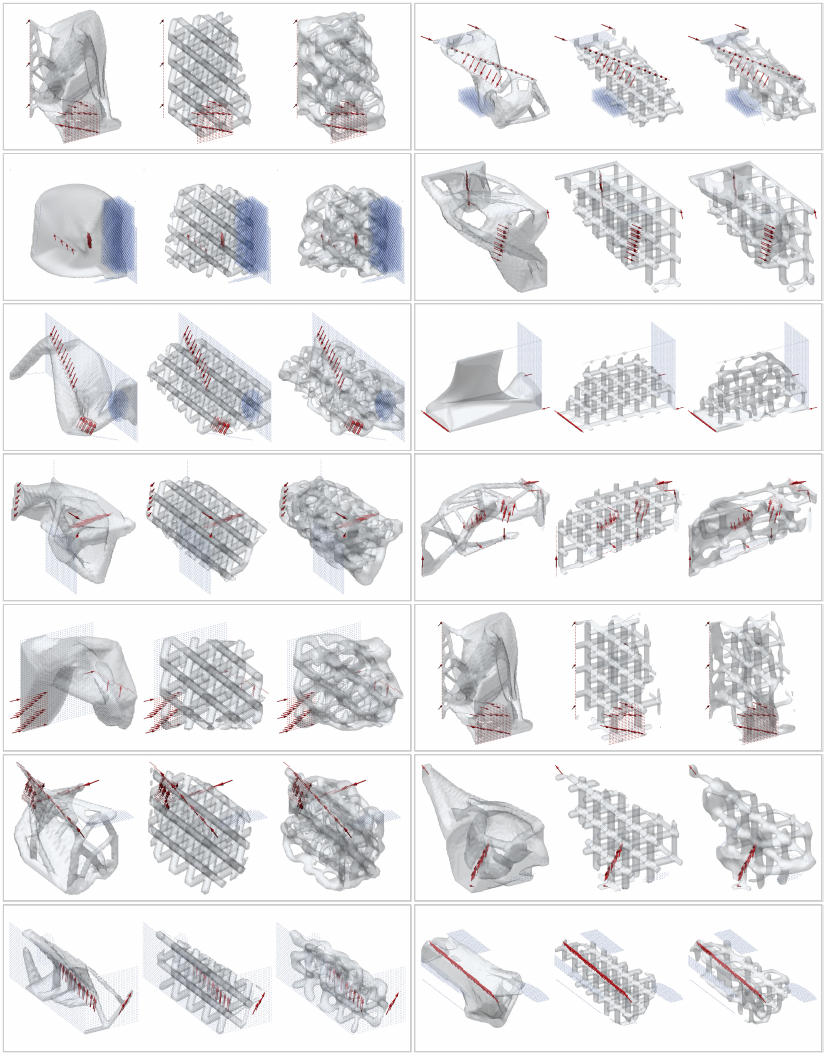}
    \caption{OAT3D on lattice redesign. Given an existing topology (left), we can use vanilla lattice fill (middle) to replace the structure in the convex hull. However, vanilla replacement can miss boundary conditions. OAT3D (right) can further improve the replaced structure through partial noising to recover the load coverage. }
    \label{fig:lattice_appendix14}
\end{figure}

\begin{figure}[H]
    \centering
    \includegraphics[width=\linewidth]{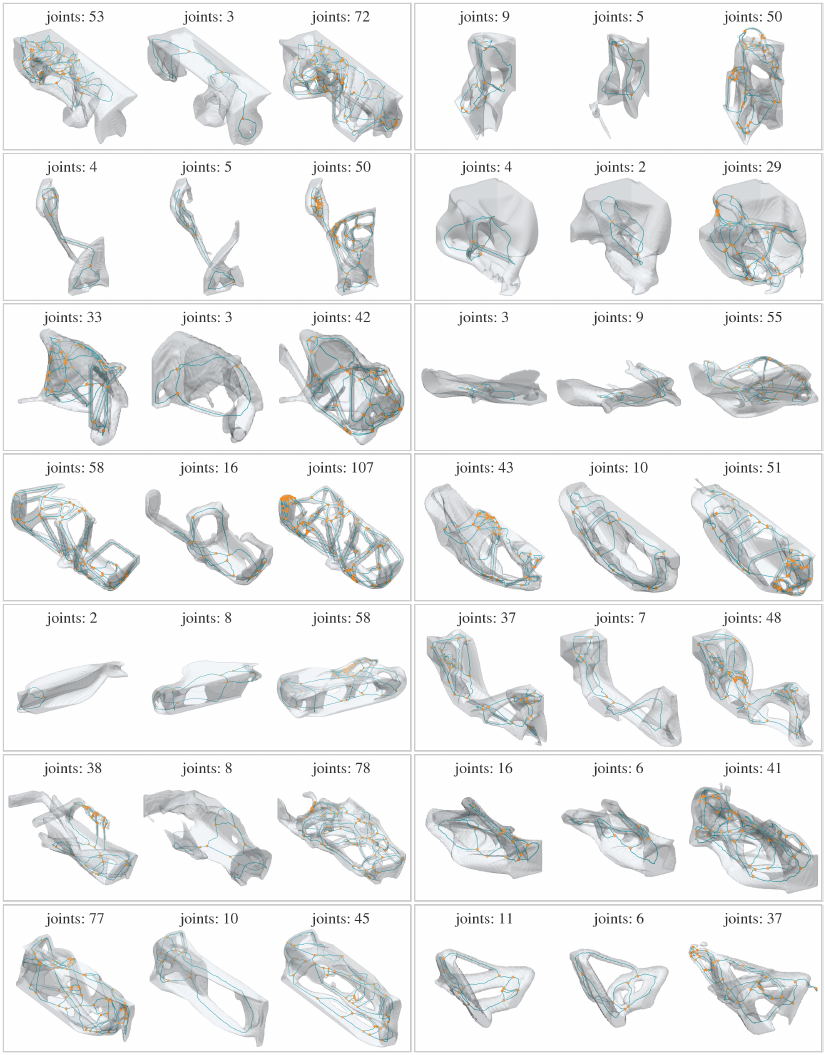}
    \caption{Joint number control with OAT3D. Joints are highlighted in orange while the skeleton is in blue. We tally up the joint number and label it on top of every topology. Left is the original topology, middle is OAT3D controlling the joint number to a lower range, while the right is OAT3D increasing the joint number.  }
    \label{fig:joint_appendix14}
\end{figure}

\newpage
\subsection{Resolution free through guided denoising}

We tested resolution-free generation using six overlapping \(128\times60\times60\) blocks, forming a \(703\times60\times60\) tower. In total, the combined structure has more than 2.5M voxels, larger than any training domain. Adjacent blocks shared 13 voxel layers, approximately 10\% of their length. Target volume fractions decreased linearly from 0.30 at the base to 0.15 at the top. Each block had a five-voxel-wide square perimeter fixed at its base and uniformly loaded in axial compression at its opposite end.

The blocks shared the same initial noise field and underwent joint denoising in each dense and sparse stage. At every step, an L1 loss measured the mean absolute difference between neighboring decoded clean estimates over the five overlapping regions. Gradients through the decoders guided both neighboring blocks toward compatibility. The resulting structure shows cyclic behavior across the blocks and remains connected, while the boundary L1 loss encourages compatibility between neighboring blocks.
\begin{figure}[H]
    \centering
    \includegraphics[width=\linewidth]{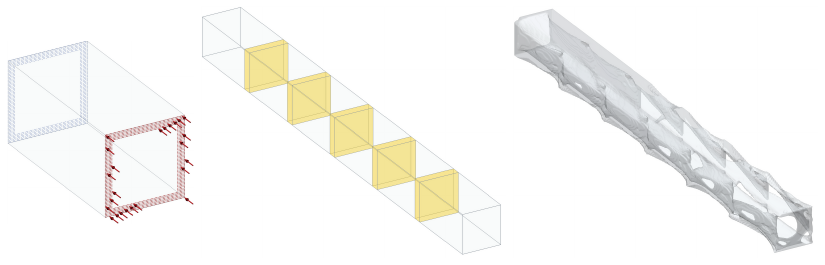}
    \caption{Joint denoising to design large scale structures. On the left is the boundary condition on a single block. The boundary condition is repeated across 6 blocks and we enforce a compatibility region between each blocks highlighted in yellow. After joint denoising, the resulting structure is shown on the right. }
    \label{fig:res_free}
\end{figure}


\begin{thebibliography}{49}
\providecommand{\natexlab}[1]{#1}
\providecommand{\url}[1]{\texttt{#1}}
\expandafter\ifx\csname urlstyle\endcsname\relax
  \providecommand{\doi}[1]{doi: #1}\else
  \providecommand{\doi}{doi: \begingroup \urlstyle{rm}\Url}\fi

\bibitem[Alzahrani et~al.(2015)Alzahrani, Choi, and Rosen]{alzahrani2015design}
Mahmoud Alzahrani, Seung-Kyum Choi, and David~W Rosen.
\newblock Design of truss-like cellular structures using relative density
  mapping method.
\newblock \emph{Materials \& Design}, 85:\penalty0 349--360, 2015.

\bibitem[Andreassen et~al.(2011)Andreassen, Clausen, Schevenels, Lazarov, and
  Sigmund]{andreassen2011efficient}
Erik Andreassen, Anders Clausen, Mattias Schevenels, Boyan~S Lazarov, and Ole
  Sigmund.
\newblock Efficient topology optimization in matlab using 88 lines of code.
\newblock \emph{Structural and Multidisciplinary Optimization}, 43\penalty0
  (1):\penalty0 1--16, 2011.

\bibitem[Baandrup et~al.(2020)Baandrup, Sigmund, Polk, and
  Aage]{baandrup2020closing}
Mads Baandrup, Ole Sigmund, Henrik Polk, and Niels Aage.
\newblock Closing the gap towards super-long suspension bridges using
  computational morphogenesis.
\newblock \emph{Nature communications}, 11\penalty0 (1):\penalty0 2735, 2020.

\bibitem[Behzadi \& Ilie{\c{s}}(2021)Behzadi and Ilie{\c{s}}]{behzadi2021real}
Mohammad~Mahdi Behzadi and Horea~T Ilie{\c{s}}.
\newblock Real-time topology optimization in 3d via deep transfer learning.
\newblock \emph{Computer-Aided Design}, 135:\penalty0 103014, 2021.

\bibitem[Bends{\o}e(1989)]{bendsoe1989optimal}
Martin~P Bends{\o}e.
\newblock Optimal shape design as a material distribution problem.
\newblock \emph{Structural optimization}, 1\penalty0 (4):\penalty0 193--202,
  1989.

\bibitem[Bends{\o}e(1995)]{bendsoe1995optimization}
Martin~P Bends{\o}e.
\newblock \emph{Optimization of structural topology, shape, and material},
  volume 414.
\newblock Springer, 1995.

\bibitem[Bends{\o}e \& Sigmund(1999)Bends{\o}e and
  Sigmund]{bendsoe1999material}
Martin~P Bends{\o}e and Ole Sigmund.
\newblock Material interpolation schemes in topology optimization.
\newblock \emph{Archive of applied mechanics}, 69\penalty0 (9):\penalty0
  635--654, 1999.

\bibitem[Bends{\o}e et~al.(2004)Bends{\o}e, Sigmund,
  et~al.]{bendsoe2004topology}
Martin~P Bends{\o}e, Ole Sigmund, et~al.
\newblock \emph{Topology optimization: theory, methods, and applications},
  volume~2.
\newblock Springer Berlin, 2004.

\bibitem[Bends{\o}e \& Kikuchi(1988)Bends{\o}e and
  Kikuchi]{bendsoe1988generating}
Martin~Philip Bends{\o}e and Noboru Kikuchi.
\newblock Generating optimal topologies in structural design using a
  homogenization method.
\newblock \emph{Computer methods in applied mechanics and engineering},
  71\penalty0 (2):\penalty0 197--224, 1988.

\bibitem[Chen et~al.(2026{\natexlab{a}})Chen, Carstensen, and
  Ahmed]{chen2026topoedit}
Hongrui Chen, Josephine~V. Carstensen, and Faez Ahmed.
\newblock {TopoEdit}: Fast post-optimization editing of topology optimized
  structures.
\newblock \emph{arXiv preprint arXiv:2602.22430}, 2026{\natexlab{a}}.
\newblock \doi{10.48550/arXiv.2602.22430}.
\newblock URL \url{https://arxiv.org/abs/2602.22430}.

\bibitem[Chen et~al.(2026{\natexlab{b}})Chen, Ha, Carstensen, and
  Ahmed]{chen2026topoctrl}
Hongrui Chen, Dat~Quoc Ha, Josephine~V. Carstensen, and Faez Ahmed.
\newblock {TopoCtrl}: Post-optimization topology editing toward target
  structural characteristics.
\newblock \emph{arXiv preprint arXiv:2603.26926}, 2026{\natexlab{b}}.
\newblock \doi{10.48550/arXiv.2603.26926}.
\newblock URL \url{https://arxiv.org/abs/2603.26926}.

\bibitem[Dittmer et~al.(2022)Dittmer, Erzmann, Harms, and
  Maass]{dittmer2022selto}
S~Dittmer, D~Erzmann, H~Harms, and P~Maass.
\newblock Selto: Sample-efficient learned topology optimization. preprint.
\newblock \emph{arXiv}, 2209, 2022.

\bibitem[Giannone \& Ahmed(2023)Giannone and Ahmed]{giannone2023topodiff}
Giorgio Giannone and Faez Ahmed.
\newblock Diffusing the optimal topology: A generative optimization approach.
\newblock In \emph{ASME 2023 International Design Engineering Technical
  Conferences and Computers and Information in Engineering Conference}, volume
  3A: 49th Design Automation Conference, pp.\  V03AT03A012. American Society of
  Mechanical Engineers, 2023.
\newblock \doi{10.1115/DETC2023-116595}.
\newblock URL \url{https://doi.org/10.1115/DETC2023-116595}.

\bibitem[Giannone et~al.(2023)Giannone, Srivastava, Winther, and
  Ahmed]{giannone2023aligning}
Giorgio Giannone, Akash Srivastava, Ole Winther, and Faez Ahmed.
\newblock Aligning optimization trajectories with diffusion models for
  constrained design generation.
\newblock \emph{Advances in neural information processing systems},
  36:\penalty0 51830--51861, 2023.

\bibitem[Groen \& Sigmund(2018)Groen and Sigmund]{groen2018homogenization}
Jeroen~P Groen and Ole Sigmund.
\newblock Homogenization-based topology optimization for high-resolution
  manufacturable microstructures.
\newblock \emph{International Journal for Numerical Methods in Engineering},
  113\penalty0 (8):\penalty0 1148--1163, 2018.

\bibitem[He et~al.(2023)He, Zhao, Lin, and Xie]{he2023hole}
Yunzhen He, Zi-Long Zhao, Xiaoshan Lin, and Yi~Min Xie.
\newblock A hole-filling based approach to controlling structural complexity in
  topology optimization.
\newblock \emph{Computer Methods in Applied Mechanics and Engineering},
  416:\penalty0 116391, 2023.

\bibitem[Herron et~al.(2024)Herron, Rade, Jignasu, Ganapathysubramanian, Balu,
  Sarkar, and Krishnamurthy]{herron2024latent}
Ethan Herron, Jaydeep Rade, Anushrut Jignasu, Baskar Ganapathysubramanian,
  Aditya Balu, Soumik Sarkar, and Adarsh Krishnamurthy.
\newblock Latent diffusion models for structural component design.
\newblock \emph{Computer-Aided Design}, 171:\penalty0 103707, 2024.

\bibitem[Ho \& Salimans(2022)Ho and Salimans]{ho2022classifier}
Jonathan Ho and Tim Salimans.
\newblock Classifier-free diffusion guidance.
\newblock \emph{arXiv preprint arXiv:2207.12598}, 2022.
\newblock \doi{10.48550/arXiv.2207.12598}.
\newblock URL \url{https://arxiv.org/abs/2207.12598}.

\bibitem[Keshavarzzadeh et~al.(2021)Keshavarzzadeh, Alirezaei, Tasdizen, and
  Kirby]{keshavarzzadeh2021image}
Vahid Keshavarzzadeh, Mitra Alirezaei, Tolga Tasdizen, and Robert~M Kirby.
\newblock Image-based multiresolution topology optimization using deep
  disjunctive normal shape model.
\newblock \emph{Computer-Aided Design}, 130:\penalty0 102947, 2021.

\bibitem[Kuszczak et~al.(2026)Kuszczak, Ku{\'s}, Bosi, and
  Bessa]{kuszczak2026meta}
Igor Kuszczak, Gawe{\l} Ku{\'s}, Federico Bosi, and Miguel~A Bessa.
\newblock Meta-neural topology optimization: knowledge infusion with
  meta-learning.
\newblock \emph{Engineering Applications of Artificial Intelligence},
  182:\penalty0 115763, 2026.

\bibitem[Lazarov et~al.(2016)Lazarov, Wang, and Sigmund]{lazarov2016length}
Boyan~S Lazarov, Fengwen Wang, and Ole Sigmund.
\newblock Length scale and manufacturability in density-based topology
  optimization.
\newblock \emph{Archive of Applied Mechanics}, 86\penalty0 (1):\penalty0
  189--218, 2016.

\bibitem[Li et~al.(2025)Li, Zou, Liu, Wang, Liang, Yu, Liu, Guo, Liang, Ouyang,
  and Cao]{li2025triposg}
Yangguang Li, Zi-Xin Zou, Zexiang Liu, Dehu Wang, Yuan Liang, Zhipeng Yu,
  Xingchao Liu, Yuan-Chen Guo, Ding Liang, Wanli Ouyang, and Yan-Pei Cao.
\newblock {TripoSG}: High-fidelity {3D} shape synthesis using large-scale
  rectified flow models.
\newblock \emph{arXiv preprint arXiv:2502.06608}, 2025.
\newblock URL \url{https://arxiv.org/abs/2502.06608}.

\bibitem[Liang et~al.(2022)Liang, Yan, and Cheng]{liang2022explicit}
Yuan Liang, XinYu Yan, and GengDong Cheng.
\newblock Explicit control of 2d and 3d structural complexity by discrete
  variable topology optimization method.
\newblock \emph{Computer Methods in Applied Mechanics and Engineering},
  389:\penalty0 114302, 2022.

\bibitem[Lipman et~al.(2023)Lipman, Chen, Ben-Hamu, Nickel, and
  Le]{lipman2023flow}
Yaron Lipman, Ricky T.~Q. Chen, Heli Ben-Hamu, Maximilian Nickel, and Matt Le.
\newblock Flow matching for generative modeling.
\newblock In \emph{The Eleventh International Conference on Learning
  Representations}, 2023.
\newblock URL \url{https://openreview.net/forum?id=PqvMRDCJT9t}.

\bibitem[Lutheran et~al.(2025)Lutheran, Das, and Tabarraei]{lutheran2025latent}
Aaron Lutheran, Srijan Das, and Alireza Tabarraei.
\newblock Latent space diffusion for topology optimization.
\newblock \emph{arXiv preprint arXiv:2508.05624}, 2025.

\bibitem[Maz{\'e} \& Ahmed(2023)Maz{\'e} and Ahmed]{maze2023diffusion}
Fran{\c{c}}ois Maz{\'e} and Faez Ahmed.
\newblock Diffusion models beat gans on topology optimization.
\newblock In \emph{Proceedings of the AAAI conference on artificial
  intelligence}, volume~37, pp.\  9108--9116, 2023.

\bibitem[Nie et~al.(2021)Nie, Lin, Jiang, and Kara]{nie2021topologygan}
Zhenguo Nie, Tong Lin, Haoliang Jiang, and Levent~Burak Kara.
\newblock Topologygan: Topology optimization using generative adversarial
  networks based on physical fields over the initial domain.
\newblock \emph{Journal of Mechanical Design}, 143\penalty0 (3):\penalty0
  031715, 2021.

\bibitem[Nobari \& Ahmed(2026)Nobari and Ahmed]{heyrani2026pgd}
Amin~Heyrani Nobari and Faez Ahmed.
\newblock Pgd-to: a scalable alternative to mma using projected gradient
  descent for multi-constraint topology optimization.
\newblock \emph{Structural and Multidisciplinary Optimization}, 69\penalty0
  (9):\penalty0 209, 2026.

\bibitem[Nobari et~al.(2024)Nobari, Regenwetter, and Ahmed]{heyrani2024towards}
Amin~Heyrani Nobari, Lyle Regenwetter, and Faez Ahmed.
\newblock Towards domain-adaptive, resolution-free 3d topology optimization
  with neural implicit fields.
\newblock In \emph{International Design Engineering Technical Conferences and
  Computers and Information in Engineering Conference}, volume 88360, pp.\
  V03AT03A012. American Society of Mechanical Engineers, 2024.

\bibitem[Nobari et~al.(2026)Nobari, Regenwetter, Picard, Han, and
  Ahmed]{heyrani2026optimize}
Amin~Heyrani Nobari, Lyle Regenwetter, Cyril Picard, Ligong Han, and Faez
  Ahmed.
\newblock Optimize any topology: a foundation model for shape-and
  resolution-free structural topology optimization.
\newblock \emph{Advances in Neural Information Processing Systems},
  38:\penalty0 153374--153419, 2026.

\bibitem[Padhi et~al.(2023)Padhi, Chakraborty, Chakrabarti, and
  Chowdhury]{padhi2023efficient}
Arya~Prakash Padhi, Souvik Chakraborty, Anupam Chakrabarti, and Rajib
  Chowdhury.
\newblock Efficient hybrid topology optimization using gpu and
  homogenization-based multigrid approach.
\newblock \emph{Engineering with Computers}, 39\penalty0 (5):\penalty0
  3593--3615, 2023.

\bibitem[Peebles \& Xie(2023)Peebles and Xie]{peebles2023scalable}
William Peebles and Saining Xie.
\newblock Scalable diffusion models with transformers.
\newblock In \emph{Proceedings of the IEEE/CVF International Conference on
  Computer Vision}, pp.\  4195--4205, 2023.
\newblock URL
  \url{https://openaccess.thecvf.com/content/ICCV2023/html/Peebles_Scalable_Diffusion_Models_with_Transformers_ICCV_2023_paper.html}.

\bibitem[Poole et~al.(2023)Poole, Jain, Barron, and
  Mildenhall]{poole2023dreamfusion}
Ben Poole, Ajay Jain, Jonathan~T. Barron, and Ben Mildenhall.
\newblock {DreamFusion}: Text-to-{3D} using {2D} diffusion.
\newblock In \emph{International Conference on Learning Representations}, 2023.
\newblock URL \url{https://arxiv.org/abs/2209.14988}.

\bibitem[Sigmund(2001)]{sigmund200199}
Ole Sigmund.
\newblock A 99 line topology optimization code written in matlab.
\newblock \emph{Structural and multidisciplinary optimization}, 21\penalty0
  (2):\penalty0 120--127, 2001.

\bibitem[Sutradhar et~al.(2010)Sutradhar, Paulino, Miller, and
  Nguyen]{sutradhar2010topological}
Alok Sutradhar, Glaucio~H Paulino, Michael~J Miller, and Tam~H Nguyen.
\newblock Topological optimization for designing patient-specific large
  craniofacial segmental bone replacements.
\newblock \emph{Proceedings of the National Academy of Sciences}, 107\penalty0
  (30):\penalty0 13222--13227, 2010.

\bibitem[Tr{\"a}ff et~al.(2023)Tr{\"a}ff, Rydahl, Karlsson, Sigmund, and
  Aage]{traff2023simple}
Erik~A Tr{\"a}ff, Anton Rydahl, Sven Karlsson, Ole Sigmund, and Niels Aage.
\newblock Simple and efficient gpu accelerated topology optimisation: Codes and
  applications.
\newblock \emph{Computer Methods in Applied Mechanics and Engineering},
  410:\penalty0 116043, 2023.

\bibitem[Wu et~al.(2017)Wu, Aage, Westermann, and Sigmund]{wu2017infill}
Jun Wu, Niels Aage, R{\"u}diger Westermann, and Ole Sigmund.
\newblock Infill optimization for additive manufacturing—approaching
  bone-like porous structures.
\newblock \emph{IEEE transactions on visualization and computer graphics},
  24\penalty0 (2):\penalty0 1127--1140, 2017.

\bibitem[Wu et~al.(2021{\natexlab{a}})Wu, Sigmund, and Groen]{wu2021topology}
Jun Wu, Ole Sigmund, and Jeroen~P Groen.
\newblock Topology optimization of multi-scale structures: a review.
\newblock \emph{Structural and Multidisciplinary Optimization}, 63\penalty0
  (3):\penalty0 1455--1480, 2021{\natexlab{a}}.

\bibitem[Wu et~al.(2021{\natexlab{b}})Wu, Xiao, and Zheng]{wu2021deepcad}
Rundi Wu, Chang Xiao, and Changxi Zheng.
\newblock {DeepCAD}: A deep generative network for computer-aided design
  models.
\newblock In \emph{Proceedings of the IEEE/CVF International Conference on
  Computer Vision}, 2021{\natexlab{b}}.
\newblock URL
  \url{https://openaccess.thecvf.com/content/ICCV2021/html/Wu_DeepCAD_A_Deep_Generative_Network_for_Computer-Aided_Design_Models_ICCV_2021_paper.html}.

\bibitem[Wu et~al.(2025)Wu, Lin, Zhang, Zeng, Yang, Bao, Qian, Zhu, Cao, Torr,
  and Yao]{wu2025direct3ds2}
Shuang Wu, Youtian Lin, Feihu Zhang, Yifei Zeng, Yikang Yang, Yajie Bao,
  Jiachen Qian, Siyu Zhu, Xun Cao, Philip Torr, and Yao Yao.
\newblock {Direct3D-S2}: Gigascale {3D} generation made easy with spatial
  sparse attention.
\newblock \emph{arXiv preprint arXiv:2505.17412}, 2025.
\newblock \doi{10.48550/arXiv.2505.17412}.
\newblock URL \url{https://arxiv.org/abs/2505.17412}.

\bibitem[Xiang et~al.(2025{\natexlab{a}})Xiang, Chen, Xu, Wang, Lv, Deng, Zhu,
  Dong, Zhao, Yuan, and Yang]{xiang2026native}
Jianfeng Xiang, Xiaoxue Chen, Sicheng Xu, Ruicheng Wang, Zelong Lv, Yu~Deng,
  Hongyuan Zhu, Yue Dong, Hao Zhao, Nicholas~Jing Yuan, and Jiaolong Yang.
\newblock Native and compact structured latents for {3D} generation.
\newblock \emph{arXiv preprint arXiv:2512.14692}, 2025{\natexlab{a}}.
\newblock URL \url{https://arxiv.org/abs/2512.14692}.

\bibitem[Xiang et~al.(2025{\natexlab{b}})Xiang, Lv, Xu, Deng, Wang, Zhang,
  Chen, Tong, and Yang]{xiang2025structured}
Jianfeng Xiang, Zelong Lv, Sicheng Xu, Yu~Deng, Ruicheng Wang, Bowen Zhang,
  Dong Chen, Xin Tong, and Jiaolong Yang.
\newblock Structured {3D} latents for scalable and versatile {3D} generation.
\newblock In \emph{Proceedings of the IEEE/CVF Conference on Computer Vision
  and Pattern Recognition}, pp.\  21469--21480, 2025{\natexlab{b}}.
\newblock URL
  \url{https://openaccess.thecvf.com/content/CVPR2025/papers/Xiang_Structured_3D_Latents_for_Scalable_and_Versatile_3D_Generation_CVPR_2025_paper.pdf}.

\bibitem[Xu et~al.(2024)Xu, Lambourne, Jayaraman, Wang, Willis, and
  Furukawa]{xu2024brepgen}
Xiang Xu, Joseph~G. Lambourne, Pradeep~Kumar Jayaraman, Zhengqing Wang, Karl
  D.~D. Willis, and Yasutaka Furukawa.
\newblock {BrepGen}: A b-rep generative diffusion model with structured latent
  geometry.
\newblock \emph{ACM Transactions on Graphics}, 2024.
\newblock URL \url{https://arxiv.org/abs/2401.15563}.

\bibitem[Yu et~al.(2019)Yu, Hur, Jung, and Jang]{yu2019deep}
Yonggyun Yu, Taeil Hur, Jaeho Jung, and In~Gwun Jang.
\newblock Deep learning for determining a near-optimal topological design
  without any iteration: Y. yu et al.
\newblock \emph{Structural and Multidisciplinary Optimization}, 59\penalty0
  (3):\penalty0 787--799, 2019.

\bibitem[Zhang et~al.(2023)Zhang, Tang, Nie{\ss}ner, and
  Wonka]{zhang2023shape2vecset}
Biao Zhang, Jiapeng Tang, Matthias Nie{\ss}ner, and Peter Wonka.
\newblock {3DShape2VecSet}: A {3D} shape representation for neural fields and
  generative diffusion models.
\newblock \emph{ACM Transactions on Graphics}, 2023.
\newblock URL \url{https://arxiv.org/abs/2301.11445}.

\bibitem[Zhang et~al.(2024)Zhang, Wang, Zhang, Qiu, Pang, Jiang, Yang, Xu, and
  Yu]{zhang2024clay}
Longwen Zhang, Ziyu Wang, Qixuan Zhang, Qiwei Qiu, Anqi Pang, Haoran Jiang, Wei
  Yang, Lan Xu, and Jingyi Yu.
\newblock {CLAY}: A controllable large-scale generative model for creating
  high-quality {3D} assets.
\newblock \emph{ACM Transactions on Graphics}, 2024.
\newblock URL \url{https://arxiv.org/abs/2406.13897}.

\bibitem[Zhang et~al.(2017)Zhang, Liu, Wei, Zhu, and Guo]{zhang2017explicit}
Weisheng Zhang, Ying Liu, Peng Wei, Yichao Zhu, and Xu~Guo.
\newblock Explicit control of structural complexity in topology optimization.
\newblock \emph{Computer Methods in Applied Mechanics and Engineering},
  324:\penalty0 149--169, 2017.

\bibitem[Zhu et~al.(2016)Zhu, Zhang, and Xia]{zhu2016topology}
Ji-Hong Zhu, Wei-Hong Zhang, and Liang Xia.
\newblock Topology optimization in aircraft and aerospace structures design.
\newblock \emph{Archives of computational methods in engineering}, 23\penalty0
  (4):\penalty0 595--622, 2016.

\bibitem[Zuo et~al.(2023)Zuo, Han, and Liu]{zuo2023explicit}
Tongxing Zuo, Haitao Han, and Zhenyu Liu.
\newblock Explicit tunnels and cavities control using simp and mma in
  structural topology optimization.
\newblock \emph{Computer-Aided Design}, 158:\penalty0 103482, 2023.

\end{thebibliography}
\end{document}